\documentclass[lettersize,journal]{IEEEtran}
\usepackage{amsmath,amsfonts}
\usepackage{array}
\usepackage{textcomp}
\usepackage{stfloats}
\usepackage{url}
\usepackage{verbatim}
\usepackage{graphicx}
\usepackage{cite}
\usepackage{algpseudocode}
\usepackage{subfigure}
\usepackage{tabularx,booktabs}
\usepackage{amssymb,amsthm}
\usepackage{xcolor}
\usepackage{multirow}
\usepackage{pifont}
\usepackage[linesnumbered,ruled]{algorithm2e}
\usepackage{soul}

\begin{document}

%\title{Blockchain-enabled Server-less\\ Federated Learning}
\title{Analysis and Evaluation of Synchronous and Asynchronous FLchain}

\author{Francesc~Wilhelmi, Lorenza~Giupponi and~Paolo~Dini % <-this % stops a space
\thanks{F. Wilhelmi (e-mail: fwilhelmi@cttc.cat) and P. Dini (e-mail: paolo.dini@cttc.es) are with Centre Tecnol\`ogic de Telecomunicacions de Catalunya (CTTC/CERCA). This work was done when L. Giupponi (lorenza.giupponi@cttc.es) was with CTTC.}% <-this % stops a space
}

\maketitle

\begin{abstract}
Motivated by the heterogeneous nature of devices participating in large-scale federated learning (FL) optimization, we focus on an asynchronous server-less FL solution empowered by blockchain technology. In contrast to mostly adopted FL approaches, which assume synchronous operation, we advocate an asynchronous method whereby model aggregation is done as clients submit their local updates. The asynchronous setting fits well with the federated optimization idea in practical large-scale settings with heterogeneous clients. Thus, it potentially leads to higher efficiency in terms of communication overhead and idle periods. To evaluate the learning completion delay of BC-enabled FL, namely \textit{FLchain}, we provide an analytical model based on batch service queue theory. Furthermore, we provide simulation results to assess the performance of both synchronous and asynchronous mechanisms. Important aspects involved in the BC-enabled FL optimization, such as the network size, link capacity, or user requirements, are put together and analyzed. As our results show, the synchronous setting leads to higher prediction accuracy than the asynchronous case. Nevertheless, asynchronous federated optimization provides much lower latency in many cases, thus becoming an appealing solution for FL when dealing with large datasets, tough timing constraints (e.g., near-real-time applications), or highly varying training data.
\end{abstract}


\maketitle

\IEEEdisplaynontitleabstractindextext

\IEEEpeerreviewmaketitle

%%%%%%%%%%%%%%%%%%%%%%%%%%%%
%% INTRODUCTION
%%%%%%%%%%%%%%%%%%%%%%%%%%%%
\section{Introduction}\label{sec:introduction}

\subsection{The rise of federated learning}

Motivated by the decentralization of machine learning (ML) training, federated learning (FL) has emerged as a popular solution by providing low communication overhead and enhanced user privacy and security~\cite{konevcny2016federated}. Different from traditional ML methods, with training data centralized in data centers, FL proposes an innovative scheme whereby training is held at end devices (also referred to as \emph{clients}) without the need for data exchange. Instead, model parameters are exchanged to optimize a global function distributively, which allows for preserving privacy among participants.

Federated optimization has opened the door to an unprecedented way of sharing insights on data among interested parties (e.g., users with sensitive information or organizations that are reluctant to disclose proprietary data), thus accelerating the proliferation of collaborative ML models with rich datasets. Accordingly, FL may represent a key enabler for efficiently and securely handling the vast quantity of information that is now used for monitoring, controlling, and optimizing operations in different sectors. Examples can be found in different areas, such as medicine~\cite{chen2020fedhealth, nguyen2021federated2} and telecommunications~\cite{lim2020federated}. In the telecom sector, the emergence of ML for networking has led to an increasing need for handling massive amounts of data~\cite{wilhelmi2020flexible}, so FL can potentially reduce network congestion while preserving privacy~\cite{niknam2020federated}. Some prominent use cases are autonomous driving~\cite{du2020federated}, unmanned aerial vehicle (UAV)-based wireless networks~\cite{brik2020federated}, edge computing~\cite{wang2019adaptive}, or Internet-of-Things (IoT) intelligence~\cite{mills2019communication}. 

\subsection{Motivation}

Despite the huge interest in FL, its mainstream setting has a number of disadvantages that prevent adoption in challenging (real-world) scenarios with thousands to millions of heterogeneous nodes. Such limitations stem from the centralized orchestration of FL, whereby a server is responsible for gathering FL local updates, aggregating them to generate a global model output, and sending the model back to clients.

To tackle the serious implications of centralization in FL (summarized in Table~\ref{tab:issues_addressed_blockchain}), blockchain has recently emerged as an appealing solution thanks to key properties like decentralization, immutability, traceability, and security~\cite{nakamoto2008bitcoin}. First, blockchain is based on distributed ledger technologies (DLT), which allows disregarding the figure of the central server to hand over the control to a decentralized network, with the positive consequences that this brings to FL (e.g., improving resilience, removing the single point of failure). As for the FL operation, the decentralization property of blockchain is very useful to mitigate the impact of stragglers~\cite{li2020federated2}, which are slow nodes that hamper the learning procedure by incurring high waiting delays. Blockchain naturally leads to an asynchronous FL operation~\cite{sprague2018asynchronous}, whereby FL devices can concurrently write and read the contents of a distributed ledger without coordination, thus reducing the iteration time while providing high accuracy~\cite{feyzmahdavian2016asynchronous, xu2021bafl}. Likewise, together with transparency and auditability, decentralized blockchains enhance democratization and ensure fairness to the FL operation.

Additional benefits of blockchain for FL are related to security and privacy. In particular, a blockchain concatenates blocks of information through advanced cryptography techniques, which allows for providing immutability~\cite{nguyen2021bedgehealth}. For instance, by combining hashing techniques (e.g., SHA-256) with digital signatures, mining mechanisms like Proof-of-Work (PoW) allow the creation of tamper-proof sequences of blocks. Thanks to the chaining of blocks through hashing, any slight change in data generates an avalanche effect that invalidates the rest of the chain. Immutability and transparency are also very important to keep track of the training procedure in FL, which enhances trustworthiness and contributes to explainability. Other unresolved security issues in traditional FL lie in data integrity and privacy. The fact is that malicious nodes can inject incorrect data to hamper the training operation~\cite{fang2020local} or can infer sensitive information from raw data exchanged during the FL procedure~\cite{mothukuri2021survey}. Blockchain can solve these issues in different manners. One prominent way stems from collective monitoring, which would allow any participant to detect and report malicious activities. In \cite{nguyen2019blockchain}, for instance, smart contracts are used for rewarding FL participants that verify the model gradients submitted to the blockchain.

Finally, another important issue in FL is the lack of incentives for participation, given that some FL devices may decide not to participate in the training operation to save computational resources or not to help competitors by sharing insights on their data. Such a lack of participation can have negative implications on the scalability of FL, so as in the achieved model's accuracy~\cite{zhan2020learning}. In this regard, blockchain provides a suitable environment to incentivize participation, as it is built upon embedded economic incentives (e.g., transaction fees) and programmable agreements~\cite{somy2019ownership}. 
Related to incentives in blockchain-enabled FL, game theory has been shown to be an appealing solution for optimizing aspects like model accuracy, completion time, or energy consumption~\cite{tu2022incentive}.

\begin{table}[ht!]
\caption{Issues in centralized FL addressed by blockchain.}
\label{tab:issues_addressed_blockchain}
\resizebox{\columnwidth}{!}{\begin{tabular}{|l|l|}
\hline
\multicolumn{1}{|c|}{\textbf{Issue in centralized FL}} & \multicolumn{1}{c|}{\textbf{Solution provided by blockchain}} \\ \hline
Straggler nodes~\cite{li2020federated2} & \begin{tabular}[c]{@{}l@{}}Decentralized and asynchronous \\ data writing/reading operations~\cite{feyzmahdavian2016asynchronous,xu2021bafl}\end{tabular} \\ \hline
\begin{tabular}[c]{@{}l@{}}Lack of democratization and \\ learning bias~\cite{kairouz2021advances} \end{tabular} & Pure decentralized control~\cite{nguyen2021federated} \\ \hline
Security bottleneck~\cite{mothukuri2021survey} & \begin{tabular}[c]{@{}l@{}}Tamper-proof properties, data \\ integrity, and strong network resilience~\cite{li2020survey}\end{tabular} \\ \hline
\begin{tabular}[c]{@{}l@{}}Participation of malicious nodes\\(e.g., local model poisoning~\cite{fang2020local})\end{tabular} & \begin{tabular}[c]{@{}l@{}}Data verification and network consensus \\ for banning malicious participants~\cite{nguyen2019blockchain}\end{tabular} \\ \hline
Privacy issues~\cite{mothukuri2021survey} & \begin{tabular}[c]{@{}l@{}}Enhanced privacy via digital signatures \\and anonymity~\cite{zhao2020privacy}\end{tabular} \\ \hline
Lack of participation~\cite{zhan2020learning} & \begin{tabular}[c]{@{}l@{}}Mechanisms for incentivizing participation \\ through rewards~\cite{somy2019ownership}\end{tabular} \\ \hline
\end{tabular}}
\end{table}

Due to the apparent benefits provided by blockchain to FL, its implementation has recently led to a new paradigm named FLchain~\cite{majeed2019flchain, bao2019flchain}, built on the foundations of decentralization, democratization, security, scalability, and trust. The introduction of blockchain to FL applications has, however, well-documented weaknesses~\cite{zheng2018blockchain}, including scalability, storage, and security aspects. Although blockchain removes the figure of the server and its associated computation and communication costs, it entails other non-negligible operational overheads (e.g., related to mining~\cite{wilhelmi2021discrete}). For that reason, the analysis and evaluation of blockchain FL are crucial for assessing the future and feasibility of FLchain solutions.

\subsection{Contributions}

To contribute to shedding light on the feasibility of FLchain, in this paper, we focus on performance and scalability aspects. To this end, we first develop an analytical model for characterizing the end-to-end FLchain latency, including the new set of delays incurred by blockchain technology. Based on the latency model, we evaluate the server-less FLchain operation and delve into the asynchronicity inherited from blockchain decentralization. The asynchronous FLchain solution (namely, \textit{a-FLchain}) is compared with a synchronous implementation of blockchain-enabled FL (namely, \textit{s-FLchain}), which can be seen as a realization of traditional FL, but without the figure of the central server. The evaluation provided in this paper allows studying the trade-off between the training time and achieved accuracy. While s-FLchain potentially achieves high accuracy, a-FLchain provides timely ML optimizations in large, heterogeneous deployments. The contributions of this paper are summarized as follows:
\begin{enumerate}
	\item We provide a tutorial-type overview of the integration of blockchain and FL technologies, namely FLchain.
	\item We develop an analytical framework to derive the blockchain latency associated with FLchain, which can be applied to both synchronous and asynchronous settings. A key aspect of the proposed framework is the batch-service queue model for measuring the queuing delay in a blockchain system, which is essential to capture asynchronous model updates.
    \item We evaluate the performance of FLchain solutions, including the synchronous (s-FLchain) and asynchronous (a-FLchain) training settings, through extensive simulations. Our focus is on the learning completion time and achieved accuracy. The analysis also serves to study the effect that blockchain parameters have on server-less FL. In particular, we adopt the EMNIST~\cite{cohen_afshar_tapson_schaik_2017} dataset to evaluate two different federated models, which range from low to high complexity.
\end{enumerate}

The rest of the paper is structured as follows: Section~\ref{section:related_work} overviews the related work on implementations and characterization of FLchain. Section~\ref{section:bc_fl} introduces FL and blockchain technologies, as well as the FLchain paradigm. Section~\ref{section:system_model} describes the system model and Section~\ref{section:delay_bc} provides the latency framework for FLchain. Section~\ref{section:results} provides numerical results of FLchain. Section~\ref{section:conclusions} concludes the paper.

%%%%%%%%%%%%%%%%%%%%%%%%%%%%
%% RELATED WORK
%%%%%%%%%%%%%%%%%%%%%%%%%%%%
\section{Related Work}
\label{section:related_work}

With the emergence of infrastructure-less network scenarios, FL optimization is moving out of the central server to be handled in a decentralized manner~\cite{lalitha2018fully,hegedHus2021decentralized}. To enable decentralized server-less FL, blockchain technology has been introduced as an appealing solution, providing trust, transparency, and immutability~\cite{liu2020secure, hou2021systematic}. In this regard, the concept of \textit{FLchain} was firstly coined in~\cite{majeed2019flchain, bao2019flchain}, where blockchain was proposed to store model updates in blocks, therefore removing the figure of the central server. FLchain has been embraced for 5G and beyond applications~\cite{lu2020blockchain2}, including use cases such as data sharing in industrial IoT~\cite{lu2019blockchain}, enabling data-driven cognitive computing in Industry 4.0~\cite{qu2020blockchained}, or enhancing trust in autonomous vehicles~\cite{pokhrel2020federated,qi2021privacy} (e.g., mobility verification, traffic prediction). %A noteworthy implementation of an FLchain simulator has been provided in~\cite{qu2021chainfl} for edge computing environments. Additionally, an Ethereum-based implementation of FLchain was provided in~\cite{korkmaz2020chain}, which was evaluated through numerical results. As for the realization of FLchain, it has been envisioned to run over multiple architectural solutions, thus being empowered by paradigms like edge computing~\cite{majeed2019flchain}, fog computing~\cite{qu2020decentralized}, or peer-to-peer (P2P) communications~\cite{kim2019blockchained, ma2020federated}. 

FLchain can be implemented both in synchronous or asynchronous modes, depending on how local models are transferred and the degree of cooperation of FL devices. In the synchronous setting, before generating a block, all the local models need to be gathered from the set of selected FL devices. This, while providing control, may entail high latency overheads, as the FL iteration time is determined by the slowest user. The straggler effect can be worsened in heterogeneous networks, where devices have different communication and computation capabilities~\cite{li2020federated}. Moreover, generating large blocks (including information from a huge number of users) contributes to the instability of the blockchain, given that forks are more likely to occur as the block propagation time increases~\cite{wilhelmi2021performance}. As a result, including many synchronized local updates in a single block (as done in~\cite{kim2019blockchained,liu2021blockchain}) may lead to severe performance issues as the number of participating FL nodes increases. Regarding asynchronous FLchain~\cite{xie2019asynchronous, xu2021asynchronous}, it has been shown to provide significant latency improvements in heterogeneous settings concerning synchronous algorithms~\cite{lian2018asynchronous}. In fact, the FL asynchronous operation fits well with FLchain implementations, given the distributed nature of blockchain systems (e.g., public blockchains running PoW). For instance, the works in~\cite{lu2020blockchain, lu2020blockchain2} proposed a PoW-based asynchronous FLchain solution for data sharing among vehicles. In this scenario, real-time communication becomes very challenging due to vehicles' unavailability and potential bandwidth constraints, so asynchronous operation is a better fit, compared to synchronous solutions.

In the literature, we find several works analyzing and evaluating various FLchain solutions. In~\cite{kim2019blockchained}, an end-to-end latency model is provided by identifying the necessary steps both for FL (compute model updates, transmit updates, and model aggregation) and blockchain (mining, validation, and block propagation). Similarly, the work in~\cite{pokhrel2020federated} follows a step analysis to derive the delay incurred by the blockchain to FL training. Another relevant work is~\cite{feng2021blockchain}, which analyzes the transaction confirmation latency of an asynchronous blockchain-enabled FL application for IoT. %Another approach that analyzes the communication cost of FLchain can be found in~\cite{lu2020low}, which is based on deep reinforcement learning.}

Different from existing works, our approach provides a more detailed blockchain delay analysis, including the queuing delay and the effect of both forks and timers, which have a notorious impact on the transaction confirmation time. Similarly to what we previously proposed in~\cite{wilhelmi2021discrete}, in this paper, we develop a novel batch-service queue model to characterize the latency of FLchain. Blockchain queue theory is an emerging field that has been recently introduced in~\cite{kawase2017transaction}, followed by the approaches in~\cite{kawase2018batch, li2018blockchain, geissler2019discrete}. The work in~\cite{kawase2017transaction} proposed a batch service queue model to understand the stochastic behavior of the transaction-confirmation process in Bitcoin. In a batch service queue, packets leave the queue in batches, rather than individually. This property is useful to capture the life-cycle of transactions in a blockchain, which are gathered before being served (mined). The batch-service queue approach has been further extended and evaluated in~\cite{kawase2018batch, li2018blockchain}. Unlike the existing literature on batch-service queuing, our model characterizes timers and forks. While timers are employed to guarantee a maximum time between mined blocks, forks are a direct consequence of the consensus protocol used by miners. As a result, disregarding the effects of timers and forks may lead to poor accuracy when deriving the blockchain queue status (e.g., the number of transactions in it) and the expected delay.

% When it comes to the analysis of asynchronous FLchain, only a reduced number of contributions are found in the literature. \textbf{CHECK: \cite{liu2021blockchain, wang2022asynchronous, xu2021bafl}.}

%%%%%%%%%%%%%%%%%%%%%%%%%%%%
%% SOLUTION PROPOSAL
%%%%%%%%%%%%%%%%%%%%%%%%%%%%
\section{Enabling Secure, Decentralized Federated Learning through Blockchain}
\label{section:bc_fl}

\subsection{Federated learning}
\label{section:federated_learning}

Formally, the goal of an FL algorithm is to solve a supervised learning optimization problem in a distributed (federated) manner. To that purpose, a set of $\mathcal{K}=\{1,2,...,K\}$ clients attempt to simultaneously optimize the global model parameters $w\in \mathbb{R}^d$ by using their local data $\mathcal{D}_k$ with size $N_k$. Clients' local losses, given by $l_k(w,\mathcal{D}_k)$, are combined to minimize a global finite-sum cost function $l(\cdot)$:
\begin{equation}
\min_{w\in \mathbb{R}^d} l(w) = \min_{w\in \mathbb{R}^d} \sum_{k=1}^{K} \frac{N_k}{N} l_k(w,\mathcal{D}_k)
\label{eq:1}
\end{equation}

To address the optimization problem in Eq.~\eqref{eq:1}, in traditional FL, a server iteratively interacts with clients that, in each iteration, provide local model updates that result from training on local (unshared) datasets. In a FL iteration, the server pushes a global model (e.g., parameters) to a set of clients, which perform on-device training (e.g., training a neural network for several epochs, based on different local data partitions) and upload their model updates. Once model updates are collected and aggregated by the server, a new global update is calculated. More iterations are run until convergence.

\begin{figure}[ht!]
    \centering
    \includegraphics[width=\linewidth]{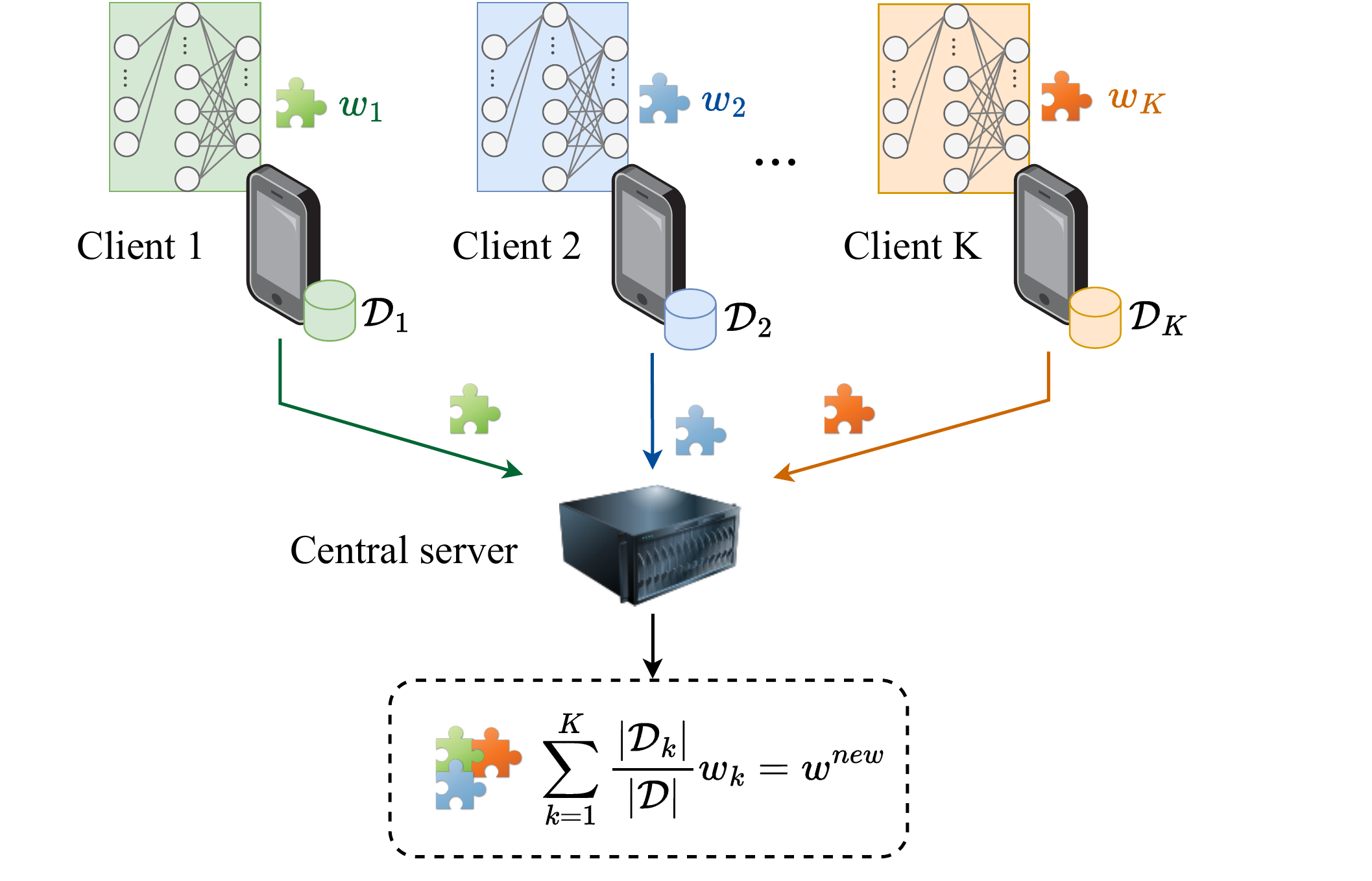}
    \caption{Model aggregation in federated learning. Clients submit local models to the server, which are then aggregated through averaging.}
    \label{fig:model_aggregation}
\end{figure}

The federated averaging (FedAvg) method~\cite{mcmahan2017communication} has been widely adopted to address FL optimization, mostly because of its effectiveness for non-convex problems and to its ability to reduce the communication overhead for synchronized stochastic gradient descent (SGD). In FedAvg, FL clients run multiple epochs of SGD before sending their locally computed gradients to the server, which updates the global model accordingly. As for the aggregation of model parameters, this is typically done by a central server (as illustrated in Fig.~\ref{fig:model_aggregation}), which collects and averages clients' updates to generate a global model. Beyond FedAvg, other optimization mechanisms like FedSGD or FedProx have been proposed to improve on convergence and efficiency aspects in FL~\cite{reddi2020adaptive}.

Another important aspect of federated optimization lies in the communication with/among clients (either for client-server or decentralized D2D architectures), as convergence depends on the acceptance of a general model over a typically high number of different local distributions. For synchronized methods to achieve consensus, assumptions such as IIDness (statistical heterogeneity on clients' data) or strong convexity of the loss function (to be twice-continuously differentiable) are typically required~\cite{chen2020convergence,li2018federated,wang2019adaptive}. However, these properties may not hold in more challenging scenarios, where datasets are heterogeneous, clients' availability is highly variable, or updates are provided asynchronously~\cite{xie2019asynchronous,chen2020asynchronous}. Under these circumstances, it is challenging to provide convergence proofs, but several works advocate for speeding up performance through complementary tools like estimating the missing local updates through DL~\cite{chen2020convergence}. 

\subsection{Blockchain technology}

Blockchain is a type of DLT that collects information in the form of transactions, which are gathered in blocks that are chained one after the other based on advanced cryptographic techniques. In essence, each block uses the hash value from the previous block to define its own structure (the first block is referred to as genesis block), thus leading to a tamper-proof sequence of information. To agree on the status of the chain (e.g., when adding a new block), a set of participating nodes (typically referred to as miners) apply certain predefined operations (e.g., solving a mathematical puzzle), commonly referred to as mining. In addition, the implementation of consensus protocols (e.g., longest chain) ensures that any malicious change on data (e.g., for double-spending purposes) would not be accepted by the majority.

\begin{figure}[ht!]
	\centering
	\includegraphics[width=\linewidth]{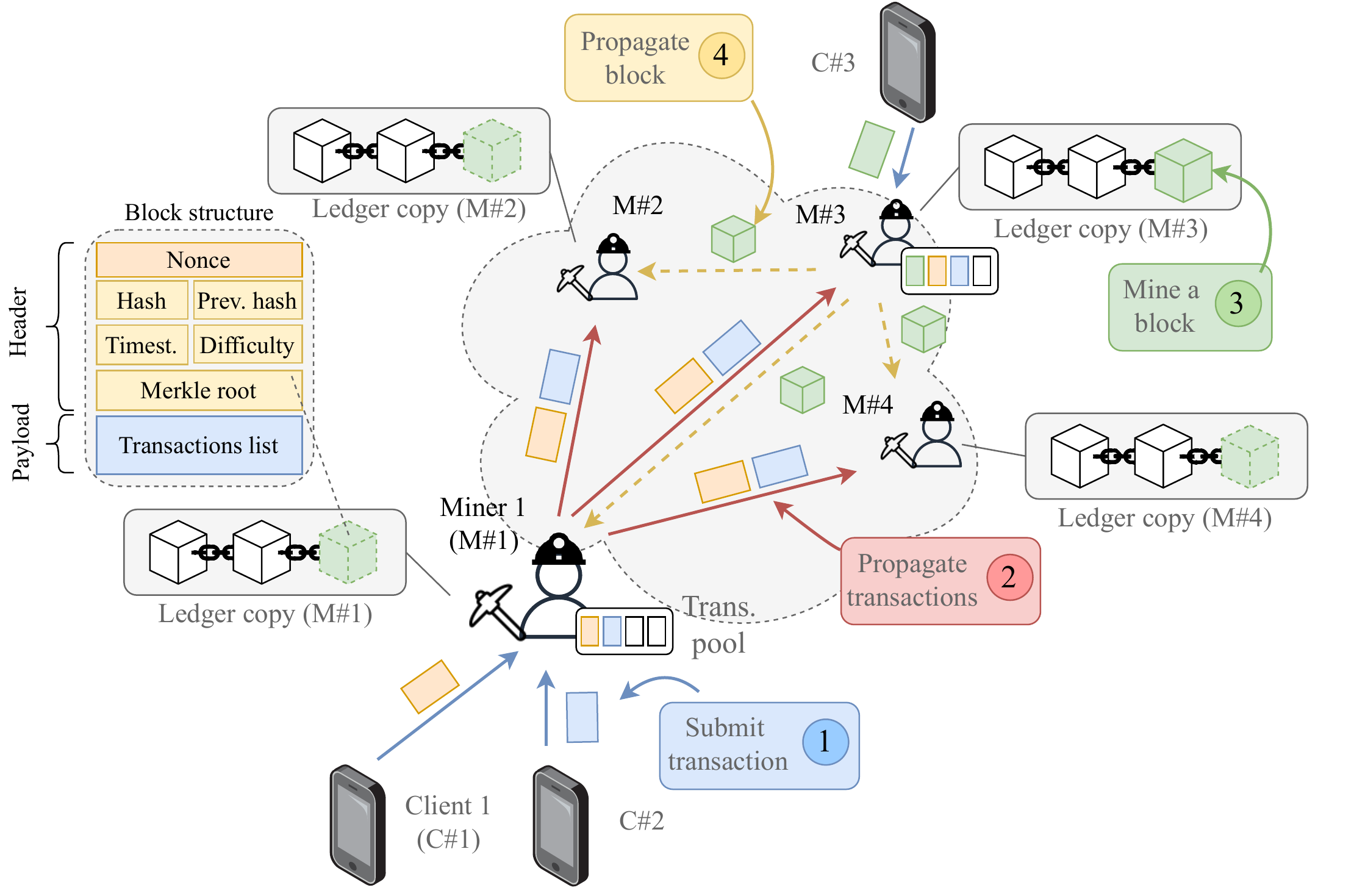}
	\caption{Overview of the procedure for mining transactions in a blockchain.}
	\label{fig:blockchain_summary}
\end{figure}

Figure~\ref{fig:blockchain_summary} summarizes the steps carried out to add transactions to a blockchain. First (\textit{point 1}, in blue), users submit transactions to the miners, which are responsible for validation and propagating them to the rest of the miners (\textit{point 2}, in red). Once a new block is ready to be appended to the blockchain, miners run a given mining algorithm for deciding the node responsible for updating the chain. In the example shown in the figure, Miner \#3 wins the mining competition and appends a block to its ledger (\textit{point 3}, in green). Finally, the newly generated block is propagated throughout the P2P blockchain network (\textit{point 4}, in yellow). Regarding the mining operation, it is important to notice that forks may occur if two or more miners come across the solution of a block at the same time, which may lead to ledger inconsistencies across miners. When a fork occurs, the longest chain prevails and invalidates the transactions included in secondary chains.

As noticed, mining and consensus are the core of blockchain technology since they allow for securely and reliably exchanging data in a decentralized (and sometimes unreliable) network. Some of the most important mining mechanisms are Proof-of-Work (PoW)~\cite{nakamoto2008bitcoin}, Proof-of-Stake (PoS)~\cite{saleh2021blockchain}, Practical Byzantine Fault Tolerance (PBFT)~\cite{castro1999practical}, or Ripple~\cite{schwartz2014ripple}. Throughout this work, we focus on PoW since it provides a high level of decentralization and raises interesting challenges in large-scale blockchain deployments.

\subsection{Blockchain for federated learning (FLchain)}
\label{section:flchain}

The realization of server-less FL through blockchain technology is often referred to as \textit{FLchain}~\cite{majeed2019flchain,nguyen2021federated}. In FLchain, a blockchain keeps track of the individual model updates 
(e.g., gradients), which allows performing model aggregation without using a central server. Notice that the global model can be built either by the clients or by some intermediary nodes (e.g., edge servers acting as miners), which allows for a fully decentralized learning procedure. In~\cite{kim2019blockchained}, for instance, the local FL model updates are included in blocks, to be later aggregated by individual devices. In contrast, in~\cite{ma2020federated}, local updates are first spread and pre-processed by miners using Gossip, and then global updates are recorded in blocks.

Taking into account the integration of blockchain into the FL operation, we identify the following iterative steps, which are executed either synchronously or asynchronously:
\begin{enumerate}
	\item \textbf{Local computation:} using local data, each client updates the model parameters by running a certain optimization mechanism (e.g., SGD).
	\item \textbf{Transactions gathering:} the local model updates generated by clients are sent to miners in the form of transactions, which are spread throughout the entire P2P network.
	\item \textbf{Block generation:} once enough valid transactions are collected (determined by the maximum block size), or if a maximum waiting time is exceeded, a candidate block is generated to be mined.
	\item \textbf{Block mining:} a mining operation (e.g., solving a puzzle) is applied to decide the next block to be appended to the blockchain.
	\item \textbf{Block propagation:} the mined block is propagated throughout the P2P network. For the sake of simplicity, we assume that blocks are simultaneously sent to all the clients once they are valid and accepted by all the miners.
	\item \textbf{Global model aggregation:} using the information included in accepted blocks (i.e., local model updates), a global model update is derived. This step could be done locally or by devices with enough computational power (e.g., edge servers). In~\cite{kim2019blockchained,ma2020federated}, for instance, each client updates the global model by itself.
	\item \textbf{Block download:} clients download the latest block from the closest miner, which contain the global model update, $w^{t+1}$.
\end{enumerate}

\begin{figure}[ht!]
	\centering
	\includegraphics[width=.8\linewidth]{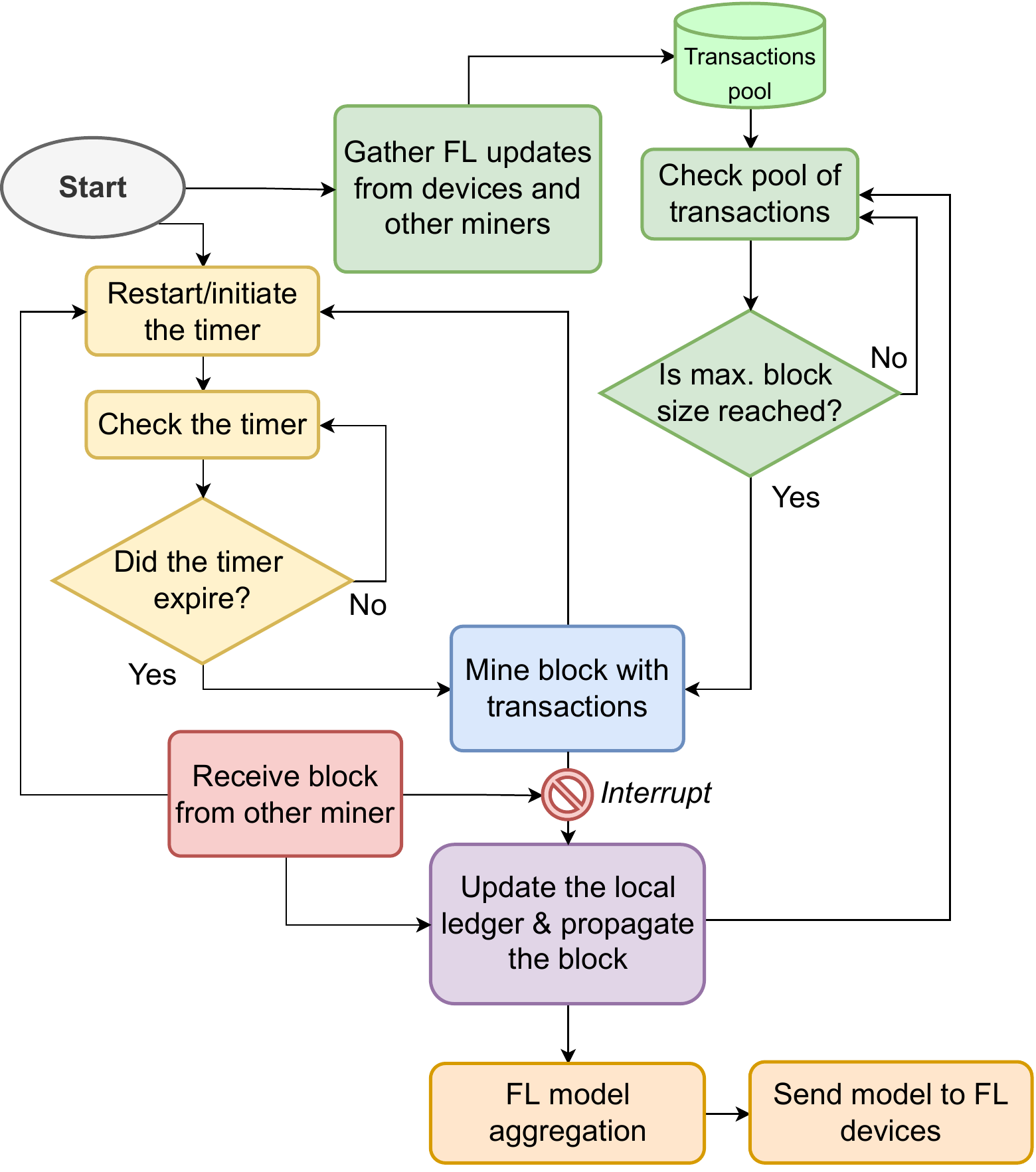}
	\caption{Flowchart of FLchain miner's operation.}
	\label{fig:flowchart_flchain_miner}
\end{figure}

The FLchain operation from the perspective of a miner is illustrated in Fig.~\ref{fig:flowchart_flchain_miner}, and applies for both synchronous and asynchronous settings. Notice that local model training is carried out by FL devices either in parallel (a-FLchain) or along with miners' operations (s-FLchain). In s-FLchain, FL devices need to wait for the latest block to start the computation of a new local update. Likewise, the block size ($S_\text{B}$) needs to be adapted in s-FLchain to accommodate all the participating clients in a single block (which contains the global model). In a-FLchain, in contrast, clients perform local computation and submit local updates asynchronously, thus working in parallel to the blockchain. A new block is generated when the maximum block size ($S_\text{B}$) is reached, or when a predefined waiting timer ($\tau$) expires.

\section{System Model}
\label{section:system_model}

\subsection{Federated learning model}
\label{section:fl_model}

Let $\mathcal{K}$ (with size~$K=|\mathcal{K}|$) be a set of clients or learners attempting to optimize a global model $w \in \mathbb{R}^d$ by minimizing a loss function $l(w)$. Each client $k\in \mathcal{K}$ has a local dataset $\mathcal{D}_k$ of size $N_k$, which is a subset of the entire dataset (of size $N$) available across all the clients $\mathcal{X} = \cup_{k=1}^K \mathcal{D}_k$. During $t\in \{1,2,...,T\}$ iterations, a subset of clients $\mathcal{K}_t \in \mathcal{K}$ with computational power $\xi_k^\text{FL}$ is sampled to compute local model updates $w_k^{t+1}, \forall k\in \mathcal{K}_t$, which are later aggregated to provide a global model update $w^{t+1}$. The FL operation stops after an arbitrary horizon or when $|l(w^t) - l(w^{t-1})| \leq \varepsilon$, being $\varepsilon \in \mathbb{R}$ a small error tolerance constant.

FedAvg with SGD is applied to compute the global model parameters. Under this setting, clients run $E$ epochs of SGD on their local data. In the traditional FL setting, local updates are sent to the server. In contrast, for FLchain, the updates are included in candidate blocks to be mined. For a single epoch, the local model parameters are updated by client $k$ as follows:
\begin{equation}
w_k^{t+1} = w^t - \eta_l \nabla l_{k}(w^t,\mathcal{D}_k),
\label{eq:2}
\end{equation}

where $\eta_l$ is the learning rate for local updates (we use $\eta$ for global updates), and $\nabla l_{k}(w^t,\mathcal{D}_k)$ is the average loss gradient of client $k$, based on the local dataset $\mathcal{D}_k$, with respect to the global model $w^t$. By aggregating the local updates $w_k^{t+1}, \forall k\in\mathcal{K}_t$ from the clients participating in an FL round, a new global model is provided as follows:
\begin{equation}
w^{t+1} = \sum_{k\in \mathcal{K}_t} \frac{N_k}{N} w_{k}^{t+1},
\label{eq:3}
\end{equation}
Notice that the contribution of each participating client is weighted by the size of its dataset, but other approaches are possible. One prominent alternative is FedProx~\cite{li2020federated}, which performs model aggregation by considering partial contributions from heterogeneous nodes. FedProx has been shown to be superior to FedAvg in the presence of stragglers.

\subsection{FLChain model}
\label{section:bc_model}

We consider a public blockchain for publishing the FL local parameters in the form of transactions, each of size $S_\text{tr}$, to be aggregated in blocks of size $S_\text{B}$ (including headers). The P2P network maintaining the blockchain is formed by a set $\mathcal{M} = \{1, 2, ..., M$\} miners, each one running PoW in parallel. Miners are considered to be at the edge of the network, so that they can collect transactions (i.e., model updates) from clients. Moreover, communication among miners is done through a mesh network. In PoW, the first miner to correctly solve and propagate the answer of a puzzle is allowed to create a new block for the blockchain, which is to be accepted by the majority of blockchain nodes. Other mechanisms in the literature~\cite{nguyen2018survey} suit a wide range of applications (e.g., public/private governance, small/big P2P networks, high/low transparency, etc.), but PoW grants a high degree of decentralization and offers robustness in massively operated blockchains.

As a result of the PoW operation, the performance of the blockchain can be hindered by forks. A fork is a split of the blockchain into different states, and can potentially compromise the overall agreement among participants, and thus threaten stability. Forks typically occur when two or more miners solve the nonce of a candidate block almost simultaneously, i.e., before the block has been propagated by a single miner. For a fixed mining difficulty ($D$) and miners' computational power ($\xi_m^\text{M}$), assuming the synchronous operation of $M$ miners and given the block propagation delay $\delta_\text{bp}$ (see the following sections for more details), we define the fork probability as:
\begin{equation}
\begin{split}
p_\text{fork} = 1 - e^{-\lambda(M-1)\delta_\text{bp}}
\end{split}
\label{eq:fork_probability}
\end{equation}

Note that the mining procedure, as widely adopted in the literature~\cite{decker2013information}, follows an exponential distribution with parameter $\lambda$ (the expected block generation time is $1/\lambda$), so the time between mined blocks can be represented by a Poisson inter-arrival process.

The synchronous and the asynchronous versions of FLchain, s-FLchain and a-FLchain, are described in detail in Algorithm~\ref{alg:sflchain} and Algorithm~\ref{alg:aflchain}, respectively.

\begin{algorithm}[ht!]	
	\SetAlgoLined
	\textbf{Initialize:} $t=1$, $\eta_l$, $\eta$, $w^t = w^0$, $\varepsilon$\\
	\While{$|l(w^t) - l(w^{t-1})| > \varepsilon$}{
		$\mathcal{K}_t \sim \mathcal{U}(\mathcal{K})$ \Comment{sample random clients from $\mathcal{K}$} \\ 
		\For{$\forall k\in \mathcal{K}_t$ \text{(in parallel)}}{
			Download $w^t$\\
			$w^{t+1}_k \leftarrow$ LocalUpdate($w^t$, $E$, $\eta_l$, $\mathcal{D}_k$)\\
			SubmitLocalUpdate($S_\text{tr}$, $C_k$)
		}
		MineBlock($\lambda$)\\
		PropagateBlock($S_\text{B}$, $C_\text{P2P}$)\\
		$w^{t+1} \leftarrow$ GlobalUpdate($w^t$, $w_k^{t+1}$, $\eta$)\\
		$ t \gets t + 1$
	}
	\textbf{LocalUpdate}($w^t$, $E$, $\eta_l$, $\mathcal{D}_k$): $E$ epochs of SGD are executed to optimize $l_k(\cdot)$\\
	\textbf{GlobalUpdate}($w^t$, $w_k^{t+1}$, $\eta$): the server aggregates the local updates $w_k^{t+1}$ to generate a global model $w^t$ \\
	\textbf{SubmitLocalUpdate}($S_\text{tr}$, $C^{}_k$): transactions with size $S_\text{tr}$ are sent through a link with capacity $C^{}_k$\\
	\textbf{MineBlock}($\lambda$): blocks are mined with block generation rate $\lambda$\\
	\textbf{PropagateBlock}($S_\text{B}$, $C_\text{P2P}$): blocks with size $S_\text{B}$ are sent through a link with capacity $C_\text{P2P}$\\
	\caption{Implementation of s-FLchain}
	\label{alg:sflchain}			
\end{algorithm}

\begin{algorithm}[ht!]	
	\SetAlgoLined
	\textbf{Initialize:} $t=1$, $\eta_l$, $\eta$, $w^t = w^0$, $\varepsilon$, $\tau$\\
	\While{$|l(w^t) - l(w^{t-1})| > \varepsilon$}{
	$u^t \leftarrow 0$
		\textbf{Clients (asynchronously):}\\
		Download $w^t$\\
		$w^{t+1}_k \leftarrow$ LocalUpdate($w^t$, $E$, $\eta_l$, $\mathcal{D}_k$)\\
		SubmitLocalUpdate($S_\text{tr}$, $C_k$)\\
		$u^t \leftarrow u^t + 1$ \\
		\textbf{Blockchain (in parallel):}\\
		\eIf{$|u^t| \geq S_\text{B}$ or $\tau$ has expired}
		{
			MineBlock($\lambda$)\\
			PropagateBlock($\min(|u_t|,S_\text{B}$), $C_\text{P2P}$)\\
			$w^{t+1} \leftarrow$ GlobalUpdate($w^t$, $w_k^{t+1}$, $\eta$)\\
			$ t \gets t + 1$\\
			Restart $\tau$
		}{
			Wait
		}
	}
	\textbf{LocalUpdate}($w^t$, $E$, $\eta_l$, $\mathcal{D}_k$): $E$ epochs of SGD are executed to optimize $l_k(\cdot)$\\
	\textbf{GlobalUpdate}($w^t$, $w_k^{t+1}$, $\eta$): the server aggregates the local updates $w_k^{t+1}$ to generate a global model $w^t$ \\
	\textbf{SubmitLocalUpdate}($S_\text{tr}$, $C^{}_k$): transactions with size $S_\text{tr}$ are sent through a link with capacity $C^{}_k$\\
	\textbf{MineBlock}($\lambda$): blocks are mined with block generation rate $\lambda$\\
	\textbf{PropagateBlock}($S_\text{B}$, $C_\text{P2P}$): blocks with size $S_\text{B}$ are sent through a link with capacity $C_\text{P2P}$\\  
	\caption{Implementation of a-FLchain}
	\label{alg:aflchain}			
\end{algorithm}

In s-FLchain, miners must wait for all the local updates before building a block. As for a-FLchain, the waiting time is affected by the transaction arrivals and the maximum waiting time $\tau$, among other parameters. In particular, we model the arrivals process through a Poisson distribution with parameter $\nu$, which is a function of the clients' activity (it depends on the size of their local datasets and their computation and communication capabilities). We define $\nu$ as follows:
\begin{equation}
\nu = \sqrt{K \Big( \text{E}[\delta_{\text{B}}^\text{DL}] + N_k\xi^\text{FL} + \text{E}[\delta_{\text{tr}}^\text{UL}] \Big)^{-1}},
\end{equation}

where $\delta_{\text{B}}^\text{DL}$ and $\delta_{\text{tr}}^\text{UL}$ are random variables characterizing the delays to download the global model in a block and to upload a local update in the form of a transaction, respectively, and $\xi^\text{FL}$ is the required amount of CPU cycles to process a single FL training data point. We would like to remark that the parameter $\nu$ is hard to estimate because it depends on the transaction confirmation latency, which also depends on $\nu$. Notice that submitting a transaction (a model update) to the blockchain is a blocking process from the point of view of a single node: once a transaction is submitted, it needs to wait until the next mined block before generating a new local update (a global model update is required before computing and submitting the next local update).

\subsection{Communication model}
\label{section:comm_model}
We consider $P$ orthogonal wireless channels under frequency division multiple access (FDMA) for supporting the communication among clients and miners. For a given transmitter-receiver pair $\{i,j\}$, the data rate of the link $R_{i,j}$, depends on the bandwidth $b_{i}$ allocated to the transmitter and the signal-to-interference-plus-noise ratio (SINR), $\gamma_{j}$, at the receiver:   
\begin{equation}
R_{i,j} = b_i \log_2(1 + \gamma_j)
\end{equation}

The SINR at node $j$ can be expressed as follows:
\begin{equation}
\gamma_j = \frac{P_{i}G_{i,j}}{\sum_{\forall l \neq i,j}P_l G_{l,j} +\sigma_0},
\end{equation}

where $P_{i}$ is the power used by transmitter $i$, $G_{i,j}$ is the channel gain, and $\sigma_0$ is the noise power. Regarding the path-loss model, the loss at distance $d$ is given by:%~\cite{feng2006path,fryziel2002path}:
\begin{equation}
PL(d) = P_t-PL_0+10 \alpha \log_{10}(d) + \frac{\sigma}{2} + \frac{d}{10} \frac{\zeta}{2},
\end{equation}

where $P_t$ is the power at the output of the transmitter's antenna, $PL_0$ is the loss at the reference distance, $\alpha$ is the path-loss exponent, $\sigma$ is the shadowing factor, and $\zeta$ is the obstacles factor. As for the links among miners in the P2P network, they are assumed to have fixed capacity.

\section{Delay Analysis of Blockchain-enabled Federated Learning}
\label{section:delay_bc}

\subsection{FLchain iteration latency}

To measure the necessary time to perform the s-FLchain and a-FLchain procedures defined in Section~\ref{section:flchain}, we identify the following delays:
\begin{enumerate}
	\item \textbf{Block filling delay ($\delta_\text{bf}$)}: a block is filled with transactions (i.e., local updates) from clients. This includes local model computation (clients run gradient descent on their own dataset to update the local model) and local model upload (clients submit their local updates to the closest miner).
	\item \textbf{Block generation delay ($\delta_\text{bg}$):} miners run PoW to find the block's nonce, or until receiving a new valid mined block.
	\item \textbf{Block propagation delay ($\delta_\text{bp}$):} mined blocks are propagated throughout the entire P2P network. We assume that all the miners receive the propagated blocks simultaneously.
	\item \textbf{Global model aggregation delay ($\delta_\text{agg}$):} a global model update is generated by aggregating the local updates from the latest mined block. The aggregation procedure can either be done by clients (fully decentralized) or by miners (edge computing-based). In case aggregation is performed by clients, the block download procedure is done first.
	\item \textbf{Block download delay ($\delta_\text{bd}$):} clients download the latest propagated block from miners, containing either the local updates from the previous iteration or the updated global model, depending on the model aggregation approach.
\end{enumerate}

Gathering all the delays in FLchain, and by taking into account the negative implications of forks, we define the expected iteration time ($\text{T}_\text{iter}$) as:
\begin{equation}
\text{T}_\text{iter} =  \frac{(\delta_\text{bf} + \delta_\text{bg} + \delta_\text{bp})}{1-p_\text{fork}} + \delta_\text{agg} + \delta_\text{bd},
\label{eq:delay}
\end{equation}

Notice that the block filling, the block generation, and the block propagation operations are affected by forks. Therefore, these steps may potentially be repeated in case conflicts arise.

\subsection{Batch-service queue model for a-FLchain}

In s-FLchain, all the devices are assumed to be synchronized, so the time for gathering transactions in a block is deterministic and depends on the slowest node:
\begin{equation}
\delta_\text{bf}^\text{sync} = \max_{k\in \mathcal{K}} (\delta_\text{c}^k + \delta_\text{ul}^k),
\end{equation}
where $\delta_\text{c}^k$ and $\delta_\text{ul}^k$ are the local computation and local model upload delays, respectively, at the $k$-th client. These delays depend on each device's computation and communication capabilities. 

\begin{figure}[ht!]
	\centering
	\includegraphics[width=\linewidth]{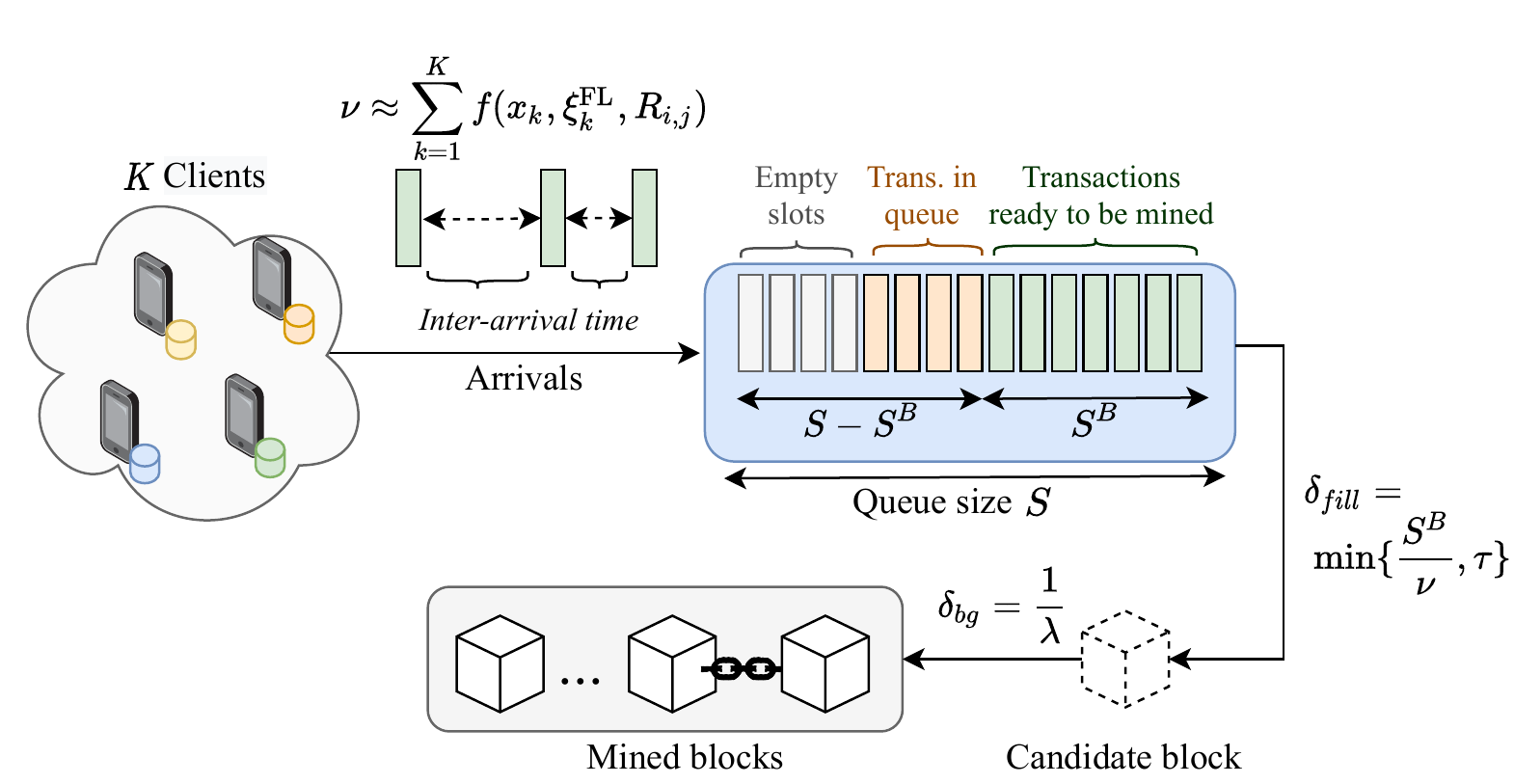}
	\caption{Batch-service queue model for a-FLchain.}
	\label{fig:batch_service_queue}
\end{figure}

In contrast, in a-FLchain, transactions are collected asynchronously, following a random distribution that depends on the clients' activity. In particular, a block is generated when enough transactions have been collected by miners (which depends on the block size $S_\text{B}$), or when a maximum waiting time $\tau$ is exceeded. To capture the block filling latency in the asynchronous case, we focus on queue theory and derive the batch-service queue model illustrated in Fig.~\ref{fig:batch_service_queue}. As shown in the figure, the block filling procedure is determined by the block size $S_\text{B}$, the arrivals rate $\nu$, and by the maximum waiting time $\tau$. If the number of arrivals is enough for filling a block before $\tau$ expires, then a block with size $S_\text{B}$ is generated following PoW. Otherwise (the timer expires), a lower number of transactions than $S_\text{B}$ is included to the candidate block. 

\begin{figure}[ht!]
	\centering
	\includegraphics[width=\columnwidth]{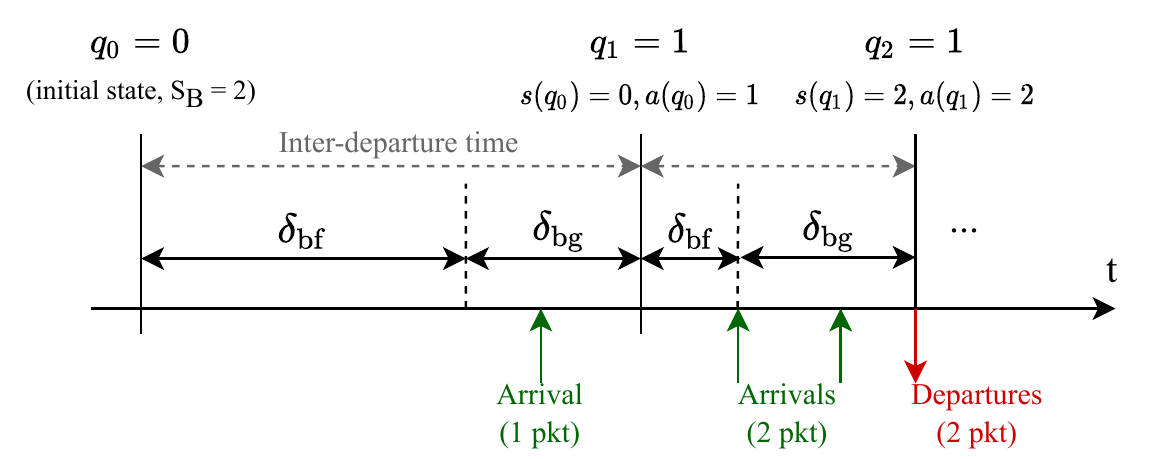}
	\caption{Inter-departure states in a batch-service queue with $S_\text{B}=2$.}
	\label{fig:queue_arrivals_example}
\end{figure}

We model the states of a finite queue (with size $S$) through a Markov chain, with states indicating the queue occupancy (number of transactions) just before a departure. In particular, the state of the batch service queue at the $n$-th departure instant is given by:
\begin{equation}
q_n = 
\min \Big\{ q_{n-1} + a(q_{n-1}) - d(q_{n-1}), S - d(q_{n-1}) \Big\},
\end{equation}
where $q_{n-1}$ is the state of the queue before the previous departure, $a(q_{n-1})$ is the number of new arrivals during the last inter-departure epoch, and $d(q_{n-1})$ is the number of packets delivered in a batch. The queue state after departures is illustrated in Fig.~\ref{fig:queue_arrivals_example}, which considers a simple example of a queue with $S_\text{B}=2$. 

As shown in the figure, the queue status at the beginning is $q_0 = 0$ (the queue is empty). During the inter-departure time, we observe an arrival, which occurred during the block generation period. In this case, $\tau$ expired before noticing any arrival during the block filling time, thus no packets are included in the mined block. During the next inter-departure period (starting from $q_1 = 1$) we find one more arrival that completes the block, so block generation is performed before the timer expires. Finally, a new arrival is noticed during the mining time, so the queue ends up in $q_2 = 1$.

The departure probability distribution $\boldsymbol{\pi}^d$ is obtained by solving $\boldsymbol{\pi}^d = \boldsymbol{\pi}^d \boldsymbol{\text{P}}$ (with normalization condition $\boldsymbol{\pi}^d \boldsymbol{1}^T = 1$), where $\text{P}$ is the transition-probability matrix for departure states. Assuming that arrivals and departures follow independent Poisson and exponential distributions, respectively, the transition probabilities in $\text{P}$, for any pair of states $(i,j)$, are obtained as follows:
\begin{equation}
\resizebox{\linewidth}{!}{%
	$p_{i,j} = \begin{cases}
	\frac{\lambda}{\lambda + \nu} \Big(\frac{\nu}{\lambda + \nu}\Big)^{j-(i-d(i))}, & [i - d(i)]^+ \leq j < S - d(i)
	\\
	\\
	1-\sum^{S-d(i)-1}_{l=0} p_{i,l}, & j = S - d(i)
	\\
	\\
	0, & \text{otherwise}
	\end{cases}$
}
\label{eq:12}
\end{equation}

The first part of Eq.~\eqref{eq:12}, $\frac{\lambda}{\lambda + \nu} \Big(\frac{\nu}{\lambda + \nu}\Big)^{j-(i-d(i))}$, indicates the probability of going from state $i$ to $j$, provided that the queue is not filled completely. This is obtained from the new distribution resulting from the combination of Poisson arrivals and exponentially distributed departures, i.e., $\int_{0}^\infty e^{-\lambda t} \frac{(\lambda t)^{j}}{j!} \cdot \mu e^{-\mu t} dt$. The second part, $1-\sum^{S-d(i)-1}_{l=0} p_{i,l}$, instead, refers to the single case where the maximum queue length is reached, which also considers arrivals exceeding the total queue size. Finally, the last case depicts the set of non-feasible transitions (e.g., $j \geq S-d(i)$).

Next, we obtain the steady-state queue distribution $\boldsymbol{\pi}^s$ by differentiating between the cases where the timer expires (with probability $\tau$) or not ($\overline{\tau}$). First, the Poisson arrivals see time averages (PASTA) property is used when the timer does not expire, i.e., when blocks are filled with transactions. The cases where the timer expires are treated differently, and require to consider the probability of observing each possible number of packets in the queue on the conditional $\tau$ time is required to prepare a block. With that, we compute $\boldsymbol{\pi}^s$ for states $s < S$ as follows:
\begin{equation}
\begin{split}
\pi^s &= \frac{1}{\nu \text{E}[T]} \sum_{i=0}^s \pi_i^d \bigg( \overline{\varsigma}_{\tau,i} \Big( \sum_{j=s-d(i)+1}^{S-d(i)} p_{i,j} \Big) \\
&+ \varsigma_{\tau,i} \Big( \sum_{j=i}^{S_\text{B}-1} \text{P}(n=j-i|\tau) \big(\sum_{l=s-d(j-i)+1}^{S-d(j-i)} p_{j,l}\big) \Big) \bigg),
\end{split}
\end{equation}
where $\text{E}[T]$ is the expected inter-departure time, $\varsigma_{\tau,i}$ is the timer expiration probability from departure state $i$ ($\overline{\varsigma}_{\tau,i} = 1 - \varsigma_{\tau,i}$), which is non-zero for $i<S_\text{B}$. The steady-state probability of finding the queue full is given by $\pi^S = 1-\sum^{S-1}_{s=0} \pi^{s}$. 

Finally, using Little's law~\cite{shortle2018fundamentals}, the expected queue delay is computed as:
\begin{equation}
\delta_\text{bf}^\text{async} = \frac{\sum_{s=0}^S s \pi^s}{\nu(1-\pi^S)}
\end{equation}

Notice, as well, that all the transactions included in the block(s) involved in forks are added twice to the blockchain. This is a worst-case scenario that has implications in the queue occupancy, thus contributing to increasing the queuing delay. Nevertheless, the consistency among the set of transactions proposed by each miner relies on the underlying communication capabilities (i.e., how transactions are propagated throughout the P2P network), on the behavior of miners (e.g., malicious miners may temporarily hold transactions to mine the next block with higher probability), and on the incentives offered by each user willing to submit a transaction (e.g., paying a fee for including a transaction in the next block).

%%%%%%%%%%%%%%%%%%%%%%%%%%%%
%% RESULTS
%%%%%%%%%%%%%%%%%%%%%%%%%%%%
\section{Performance Evaluation}
\label{section:results}

In this section, we evaluate the performance of FLchain and provide insights on its suitability.\footnote{All the source code used in this project is open-access and can be accessed at \url{www.github.com/fwilhelmi/blockchain_enabled_federated_learning}, commit: 648ead5. Accessed: July 29, 2022.} First, we analyze aspects of the blockchain and assess the sensitivity of this technology on various important parameters, such as the arrivals rate ($\nu$), the block size ($S_\text{B}$), or the block generation rate ($\lambda$). Then, we evaluate both s-FLchain and a-FLchain solutions. The simulation parameters are collected in Table~\ref{tab:sim_parameters}.

\begin{table}[ht!]
	\centering
	\caption{Simulation parameters.}
	\label{tab:sim_parameters}
	\resizebox{\columnwidth}{!}{\begin{tabular}{|c|c|l|c|}
			\hline
			\multicolumn{1}{|l|}{} & \textbf{Parameter} & \multicolumn{1}{c|}{\textbf{Description}} & \multicolumn{1}{c|}{\textbf{Value}} \\ \hline
			\multirow{5}{*}{\rotatebox{90}{\textbf{Blockchain}}} & $S_\text{tr}$ & Transaction size & 5 Kbits \\ \cline{2-4} 
			& $S_{h}$ & Block header size & 200 Kbits \\ \cline{2-4} 
			& $M$ & Number of miners & 10 \\ \cline{2-4} 
			& $\tau$ & Max. waiting time & 1,000 s \\ \cline{2-4} 
			& $S$ & Queue length & 1,000 \\ \hline
			% & $\xi^{M}$ & Mining computational power &
			\multirow{10}{*}{\rotatebox{90}{\textbf{Communication}}} & $d_{\min}/d_{\max}$ & Min/max distance client-BS & 0 / 4.15 m \\ \cline{2-4} 
			& $b$ & Bandwidth & 180 kHz \\ \cline{2-4} 
			& $F_c$ & Carrier frequency & 2 GHz \\ \cline{2-4} 
			& $G$ & Antenna gain & 0 dBi \\ \cline{2-4} 
			& $P_t$ & Transmit power & 20 dBm \\ \cline{2-4} 
			& $PL_0$ & Loss at the reference dist. & 5 dB \\ \cline{2-4} 
			& $\alpha$ & Path-loss exponent & 4.4 \\ \cline{2-4} 
			& $\sigma$ & Shadowing factor & 9.5 \\ \cline{2-4} 
			& $\zeta$ & Obstacles factor & 30 \\ \cline{2-4}
			& $\sigma_0$ & Ground noise & -95 dBm \\ \cline{2-4}
			& $C_\text{P2P}$ & Capacity P2P links & 5 Mbps \\ \hline
			\multirow{10}{*}{\rotatebox{90}{\textbf{Fed. Learning}}} & $\mathcal{A}$ & Learning algorithm & Neural Network \\ \cline{2-4} 
			& $h_l$ & Number of hidden layers & 2 \\ \cline{2-4} 
			& $a$ & Activation function & ReLU \\ \cline{2-4} & $o$ & Optimizer & SGD \\ \cline{2-4}
			& $l$ & Loss function & Cat. Cross-Entropy \\ \cline{2-4}
			& $\eta_l$/$\eta$ & Learning rate (local/global) & 0.01/1 \\ \cline{2-4}
			& $E$ & Epochs number & 5 \\ \cline{2-4}
			& $B$ & Batch size & 20 \\ \cline{2-4}
			& $\xi^\text{FL}$ & CPU cycles to process a data point & $10^{-5}$ \\ \cline{2-4}
			& $\xi^\text{FL}_k$ & Clients' clock speed & 1~GHz \\ \hline
		\end{tabular}
	}
\end{table}

\subsection{Blockchain queue delay}

\begin{figure}[ht!]
	\centering
	\includegraphics[width=.65\linewidth]{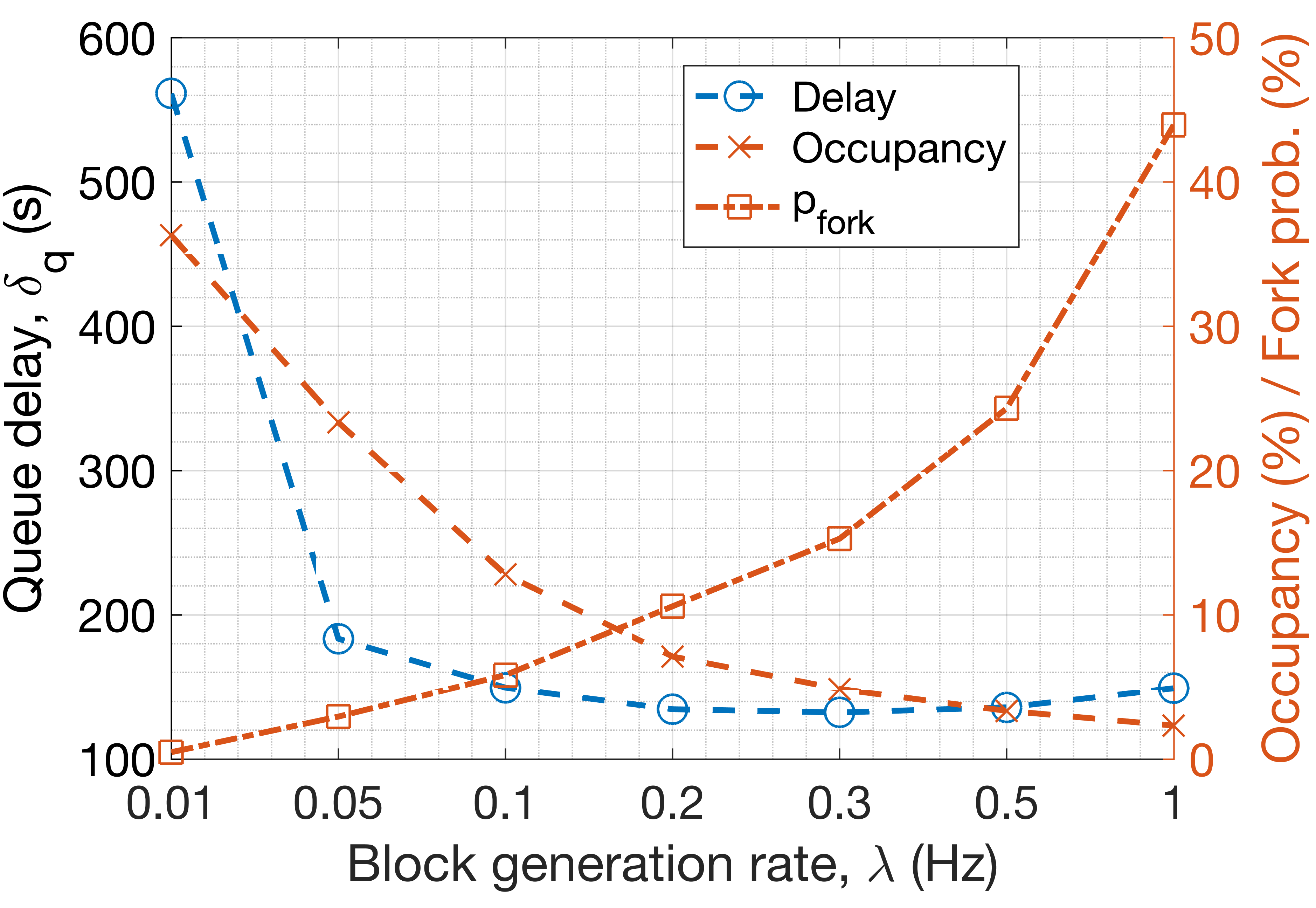}
	\caption{Mean blockchain queuing delay as a function of $\lambda$. The queue occupancy and the fork probability ($p_\text{fork}$) are also plotted. Different arrival rates $\nu$ and block sizes $S_\text{B}$ are considered and averaged in each data point.}
	\label{fig:bc_delay_mu}
\end{figure}

Fig.~\ref{fig:bc_delay_mu} shows the average queue performance, including the expected delay, average occupancy, and fork probability, for different block generation rates, block sizes, and arrival rates. As expected, the queue occupancy decreases as $\lambda$ increases, which allows processing transactions faster. However, the fork probability also increases with $\lambda$, which may compromise the performance of the blockchain.

\begin{figure}[ht!]
	\centering
	\includegraphics[width=.7\linewidth]{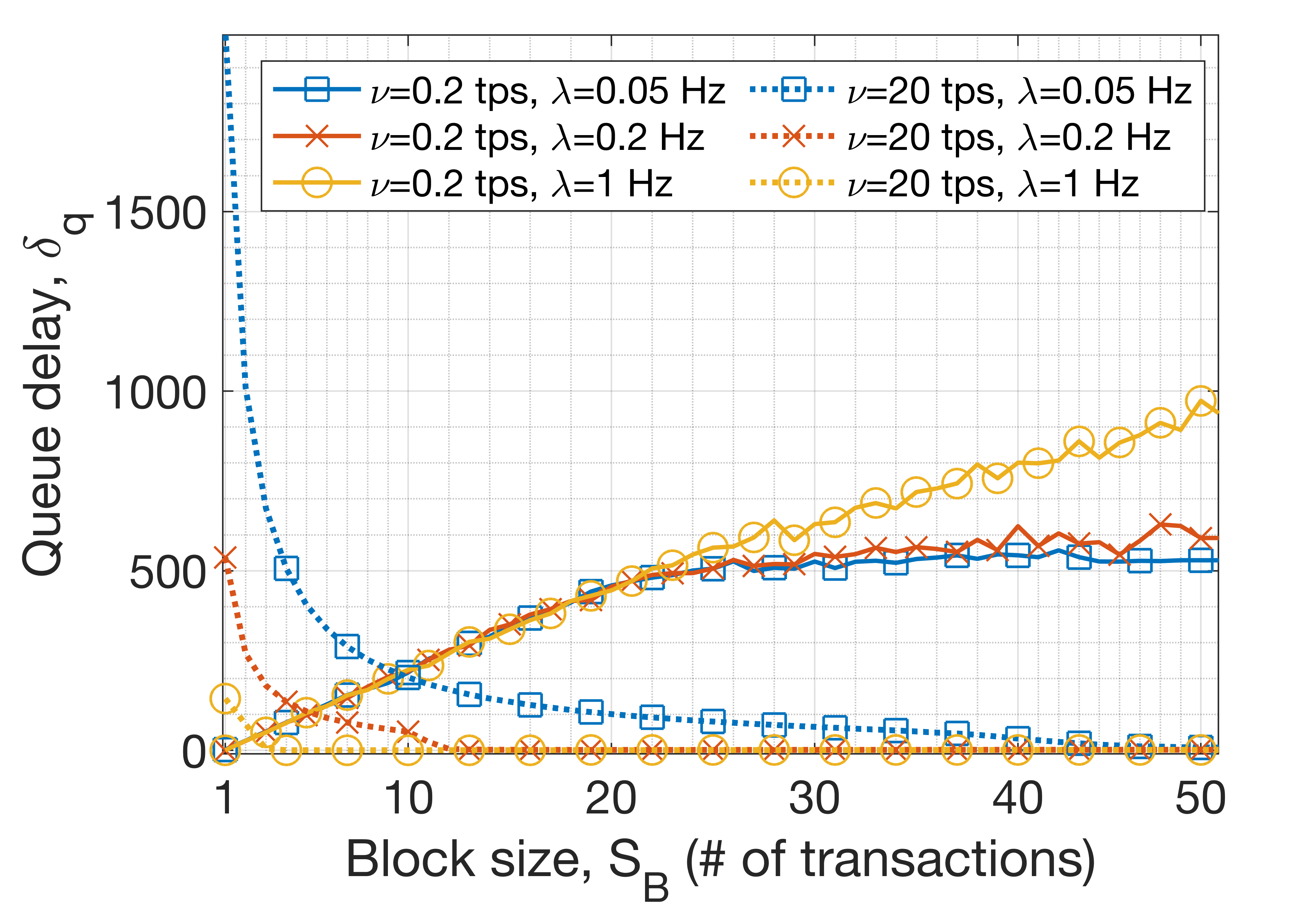}
	\caption{Blockchain queue delay as a function of the block size ($S_\text{B}$). Different $\nu$ (in transactions per second) and $\lambda$ values are considered.}
	\label{fig:2_block_size_lambda02}
\end{figure}

Alternatively, Fig.~\ref{fig:2_block_size_lambda02} analyzes the impact of the block size on the queuing delay, for representative $\nu$ (low and high traffic, respectively) and $\lambda$ values ($\{0.05, 0.2, 1\}$ Hz). As shown in the figure, the behavior of the queue delay differs for different $\nu$ values. On the one hand, for a high arrivals rate (e.g., $\nu = 20$), the queue delay is very high for short block sizes, as the queue is filled with transactions that cannot be delivered in time. The effect is the opposite for a few arrivals ($\nu = 0.2$), where the queue delay increases with the block size because the queued transactions need to wait until a block is filled.

\subsection{Blockchain transaction confirmation latency}

\begin{figure}[ht!]
	\centering
	\includegraphics[width=.65\linewidth]{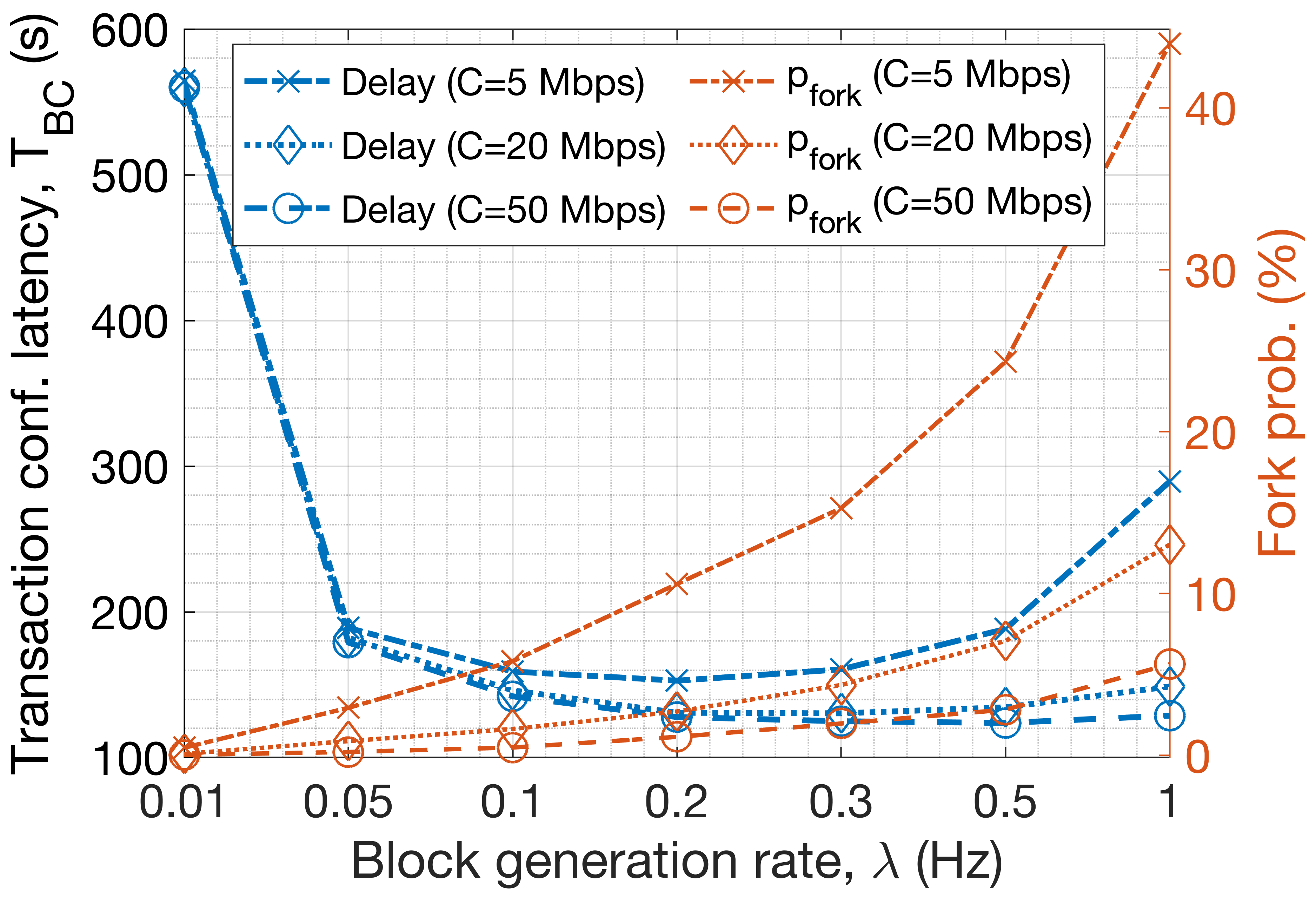}
	\caption{Blockchain transaction confirmation latency ($\text{T}_\text{BC}$) and fork probability as a function of the block generation rate ($\lambda$).}
	\label{fig:4_transaction_confirmation_delay}
\end{figure}

We focus on the transaction confirmation latency, which includes transmission times (transaction and block propagation) and the re-transmissions caused by forks, as captured in \eqref{eq:delay}. Fig.~\ref{fig:4_transaction_confirmation_delay} illustrates the blockchain's average transaction confirmation latency ($\text{T}_\text{BC}$), together with the fork probability, for different block generation rates. In addition, different blockchain P2P network capacities have been displayed, including capacities of $C_\text{P2P}=\{5, 20, 50\}$~Mbps.

Similar to the queue delay in Fig.~\ref{fig:bc_delay_mu}, the blockchain transaction confirmation latency has a concave shape, but now the effect of forks is exacerbated (especially for low capacities). The fact is that transactions involved in forks need to be re-included in the queue, which becomes critical as $\lambda$ increases. In contrast, high $C_\text{P2P}$ values allow mitigating the effect of forks and reducing the transaction confirmation latency.

\begin{figure}[ht!]
	\centering
	\includegraphics[width=.8\linewidth]{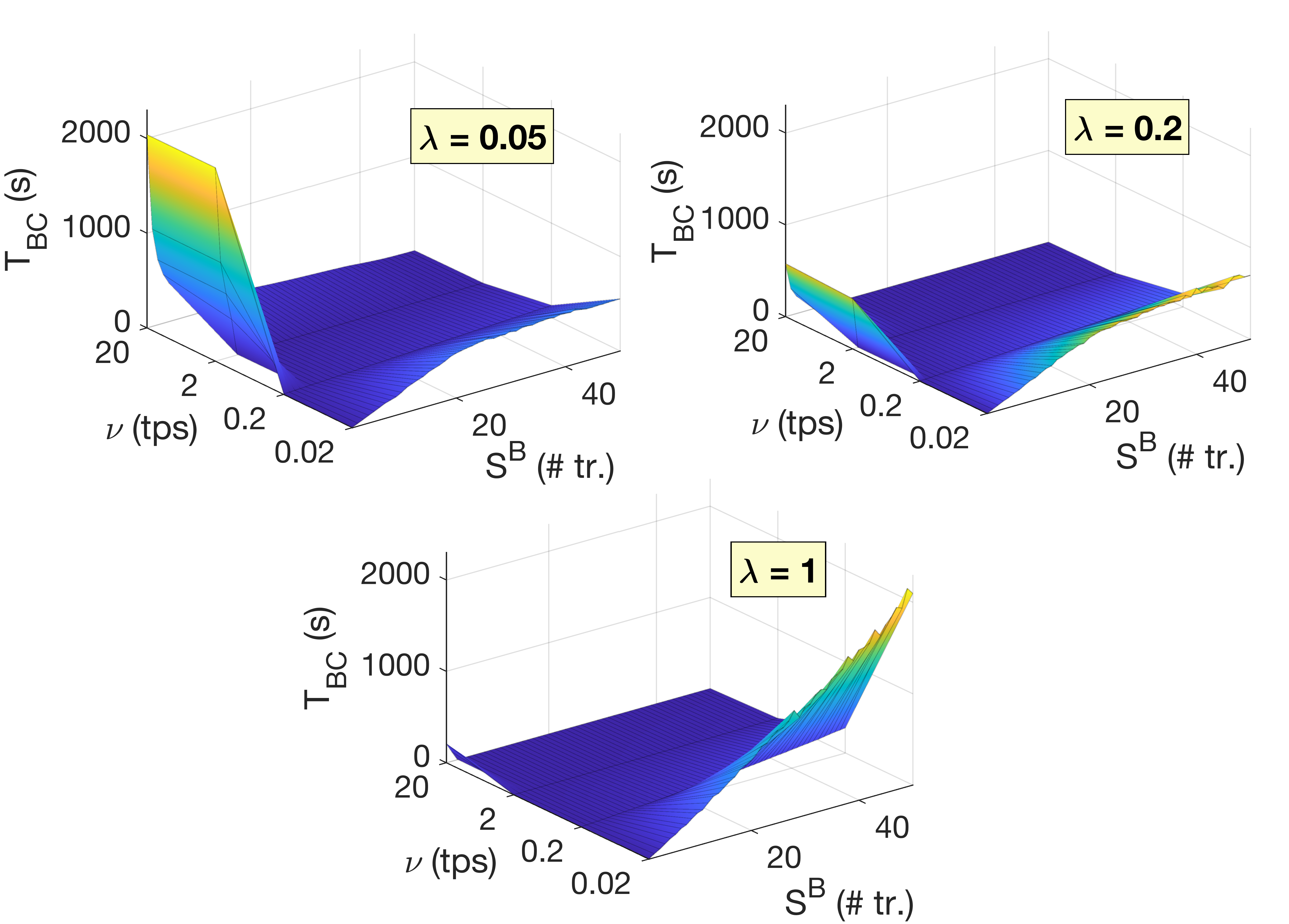}
	\caption{Blockchain transaction confirmation latency ($\text{T}_\text{BC}$) as a function of the block size ($S_\text{B}$) and the arrivals rate ($\nu$), for $\lambda = \{0.05, 0.2, 1\}$ Hz.}
	\label{fig:5_delay_vs_block_size}
\end{figure}

Finally, Fig.~\ref{fig:5_delay_vs_block_size} plots the transaction confirmation latency as a function of the block size and the arrivals rate. In particular, we show the results obtained by using $\lambda = \{0.05, 0.2, 1\}$~Hz with $C_\text{P2P} = 5$~Mbps.

Again, the relationship between the transaction confirmation latency and the block size behaves differently for different block generation ratios. For a low mining capacity (i.e., $\lambda=0.05$~Hz), using a small block size leads to very high delays, so that queue overflow occurs when heavy loads (e.g., $\nu=\{2,20\}$ transactions per second) cannot be absorbed. In contrast, for higher $\lambda$ values, the behavior varies and the trade-off between the block size and the fork probability becomes more evident. In any case, the proper election of the block size is essential for the sake of efficiency.

\subsection{s-FLchain vs a-FLchain}
\label{section:flchain_analysis}

To evaluate the s-FLchain and a-FLchain solutions, we use a popular dataset for federated image classification: \mbox{EMNIST}~\cite{cohen_afshar_tapson_schaik_2017}. The EMNIST dataset contains images of hand-written digits, distributed across 3,383 different users. More specifically, it includes 341,873 training examples and 40,832 test samples for evaluation. We solve the classification task using two different deep learning models. In such a way, we characterize two classes of problems, ranging from low to high complexity. For that purpose, we use a simple out-of-the-box feed-forward neural network (FNN) and a more complex convolutional neural network (CNN)~\cite{he2016deep} to solve the handwritten image recognition problem posed by EMNIST. These two models, as a result of their size,\footnote{The proposed FNN and CNN models have 203,530 and 2,374,506 parameters, respectively. Considering that each parameter can be represented by a 2-bytes \textit{int}, their corresponding size is 0.407 MB and 4.749 MB.} are expected to be impacted differently by the blockchain. Table~\ref{tbl:model_details} provides a summary of the models' architecture.

\begin{table}[ht!]
\caption{Summary of models' architecture.}
\label{tbl:model_details}
\begin{tabular}{|c|c|c|c|}
\hline
\textbf{Model} & \multicolumn{1}{c|}{\textbf{Layer (activation)}} & \textbf{Output shape} & \textbf{\# of param.} \\ \hline
\multirow{3}{*}{FNN} & Input & 784 & 0 \\ \cline{2-4} 
 & Dense (ReLU) & 256 & 299,969 \\ \cline{2-4} 
 & Output (Softmax) & 10 & 2,570 \\ \hline
\multirow{5}{*}{CNN} & Input & 784 & 0 \\ \cline{2-4} 
 & Conv2D (ReLU) & (26, 26, 32) & 320 \\ \cline{2-4} 
 & Conv2D (ReLU) & (24, 24, 32) & 9,248 \\ \cline{2-4} 
 & Dense (ReLU) & 512 & 2,359,808 \\ \cline{2-4} 
 & Output (Softmax) & 10 & 5,130 \\ \hline
\end{tabular}
\end{table}

In our evaluation, we split the dataset into several partitions to focus on a total of $K=\{10, 50, 100, 200\}$ random clients. Furthermore, to assess the impact of the asynchronous method, we adjust the block size in each FL round to consider smaller subsets of users. So, we define $\Upsilon$ as the percentage of users required to construct a block, assuming that all the local updates have the same size. Notice that the synchronized case corresponds to $\Upsilon=100\%$ since all the considered clients' updates are included in a single block. The timer $\tau$ is set to an arbitrarily high value to disregard its effect and focus on the block size only. 

Global model evaluation is carried out throughout each FL round at $K_\text{eval} = 50$ different clients, which are independent of $K$ and are randomly sampled from the entire dataset in each iteration. Furthermore, we use two variations of the EMNIST dataset, corresponding to independent and identically distributed (IID) and non-IID properties. Clients typically have a rich number of samples for all the classes in the original dataset~\cite{lu2020blockchain}, so we break such an IIDness and restrict each client dataset to 3 classes only (uniformly selected at random) in the non-IID case. The IID case includes all the original samples in each client, containing data from up to 10 classes. In summary, each client has an average of 101.02 (30.32 in the non-IID setting) and 12.06 images for training and evaluation, respectively.

\begin{figure}[ht!]
	\centering
    \subfigure[]{\includegraphics[width=.35\textwidth]{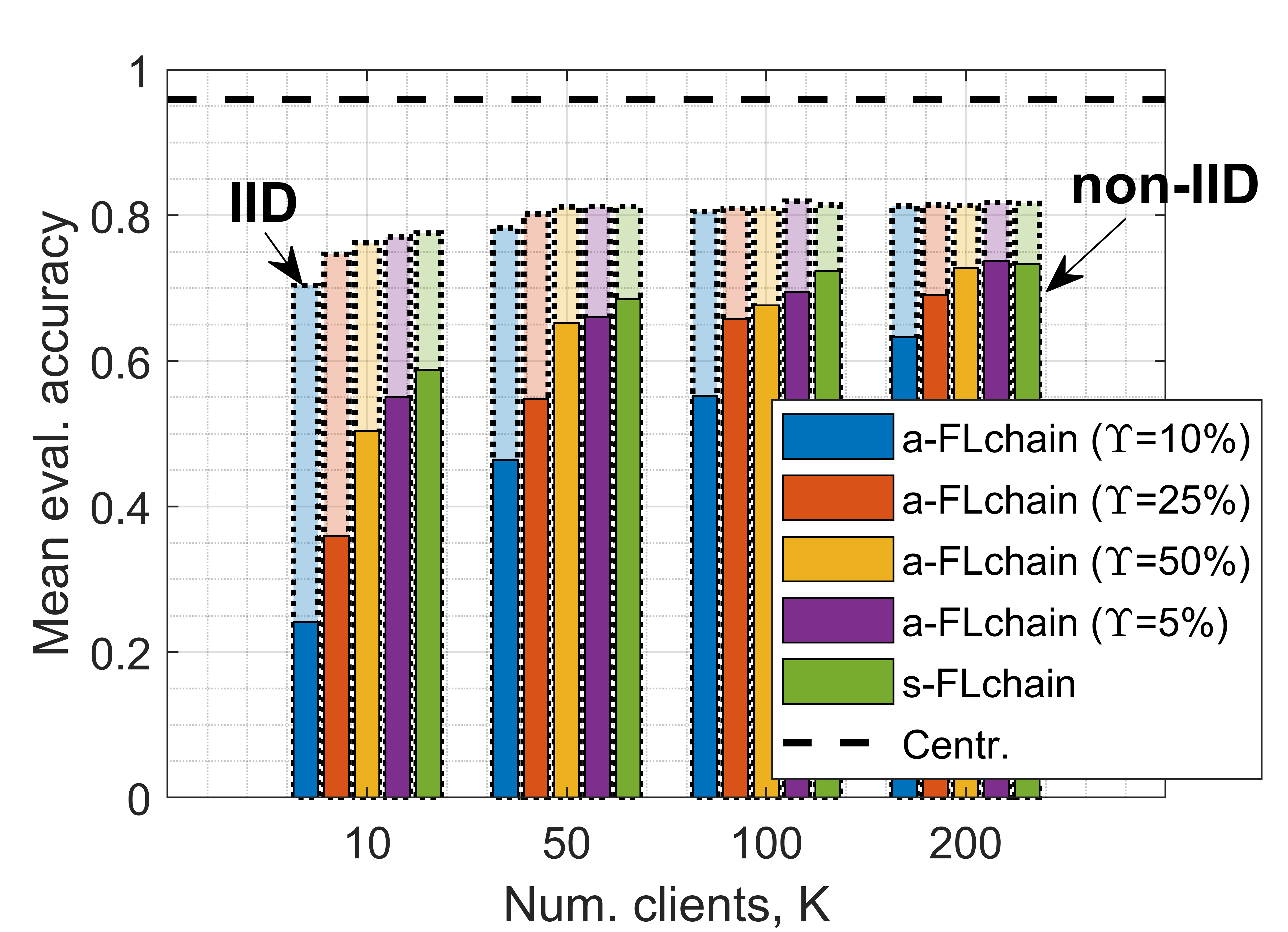}\label{fig:1_evaluation_accuracy_fnn}}
	\subfigure[]{\includegraphics[width=.35\textwidth]{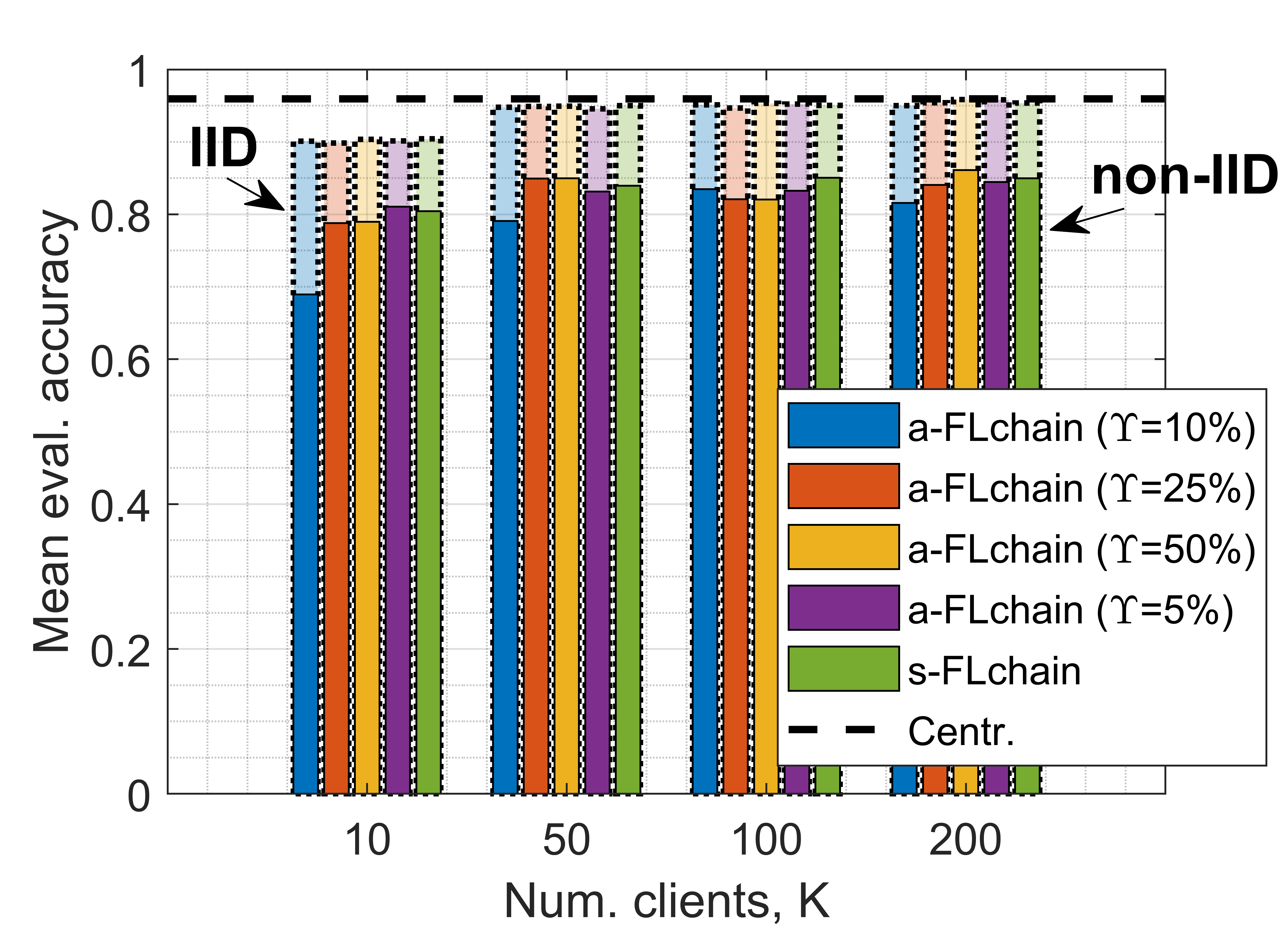}\label{fig:1_evaluation_accuracy_cnn}}
	\caption{Mean evaluation accuracy obtained by FNN and CNN models during the last 50 of 200 FL rounds on EMNIST: (a) FNN, (b) CNN. The IID (dashed bars) and non-IID (solid bars) versions of the EMNIST dataset are evaluated for a different number of clients $K$ and client sub-sampling percentages~$\Upsilon$. The results of the centralized counterpart of each mechanism are represented by a black dashed line.}
	\label{fig:1_evaluation_accuracy}
\end{figure}

Fig.~\ref{fig:1_evaluation_accuracy} shows the mean evaluation accuracy achieved by the FNN and CNN models throughout 200 FL rounds, for both s-FLchain and a-FLchain settings, and IID and non-IID datasets distributions. Furthermore, we provide the baseline results obtained by the centralized counterparts of FNN and CNN, trained using the entire EMNIST dataset (i.e., taking samples from all the clients) in a single location. The centralized counterpart of each model uses the same architecture and hyper-parameters as in the federated settings. However, it is important to notice that the centralized setting is expected to provide higher accuracy than the federated one as a result of using all the examples in the dataset for training. In FL, in contrast, a smaller number of clients is sampled overall.

As shown in Fig.~\ref{fig:1_evaluation_accuracy_fnn}, the FNN achieves an acceptable accuracy for all the block sizes (represented by $\Upsilon$) in the IID setting, which is improved as $K$ increases. However, in the non-IID case, the performance of FNN drops dramatically, especially for low $K$ and $\Upsilon$ values (e.g., for $K=10$ and $\Upsilon=10\%$). As for the performance achieved by the CNN (see Fig.~\ref{fig:1_evaluation_accuracy_cnn}), it is very close to the centralized baseline when data is IID. Nevertheless, the performance drops in the non-IID setting are not as meaningful as for the FNN model.

\begin{figure*}[ht!]
	\centering
	\subfigure[]{\includegraphics[width=.45\textwidth]{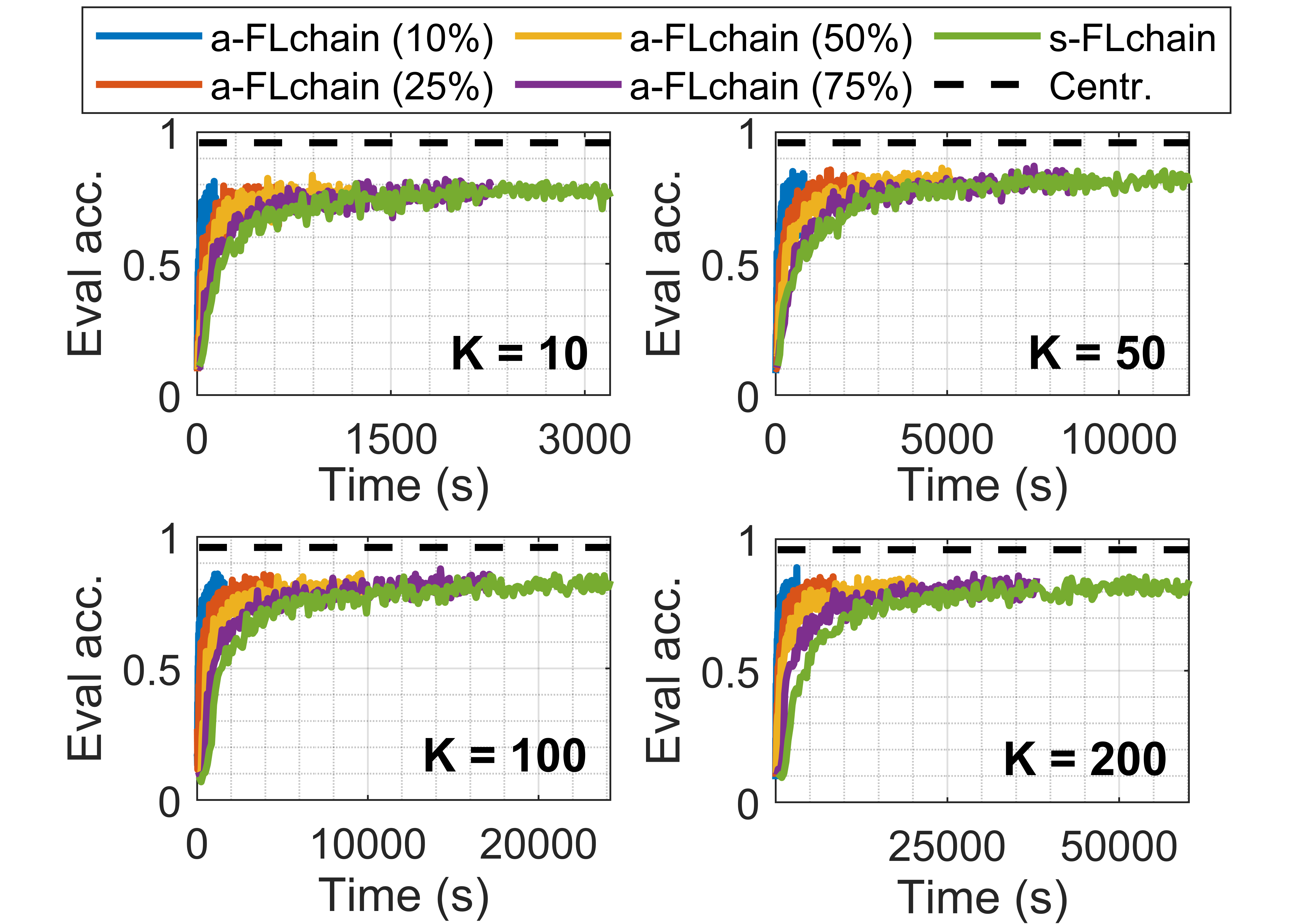}\label{fig:temporal_accuracy_emnist_fnn}}
	\subfigure[]{\includegraphics[width=.45\textwidth]{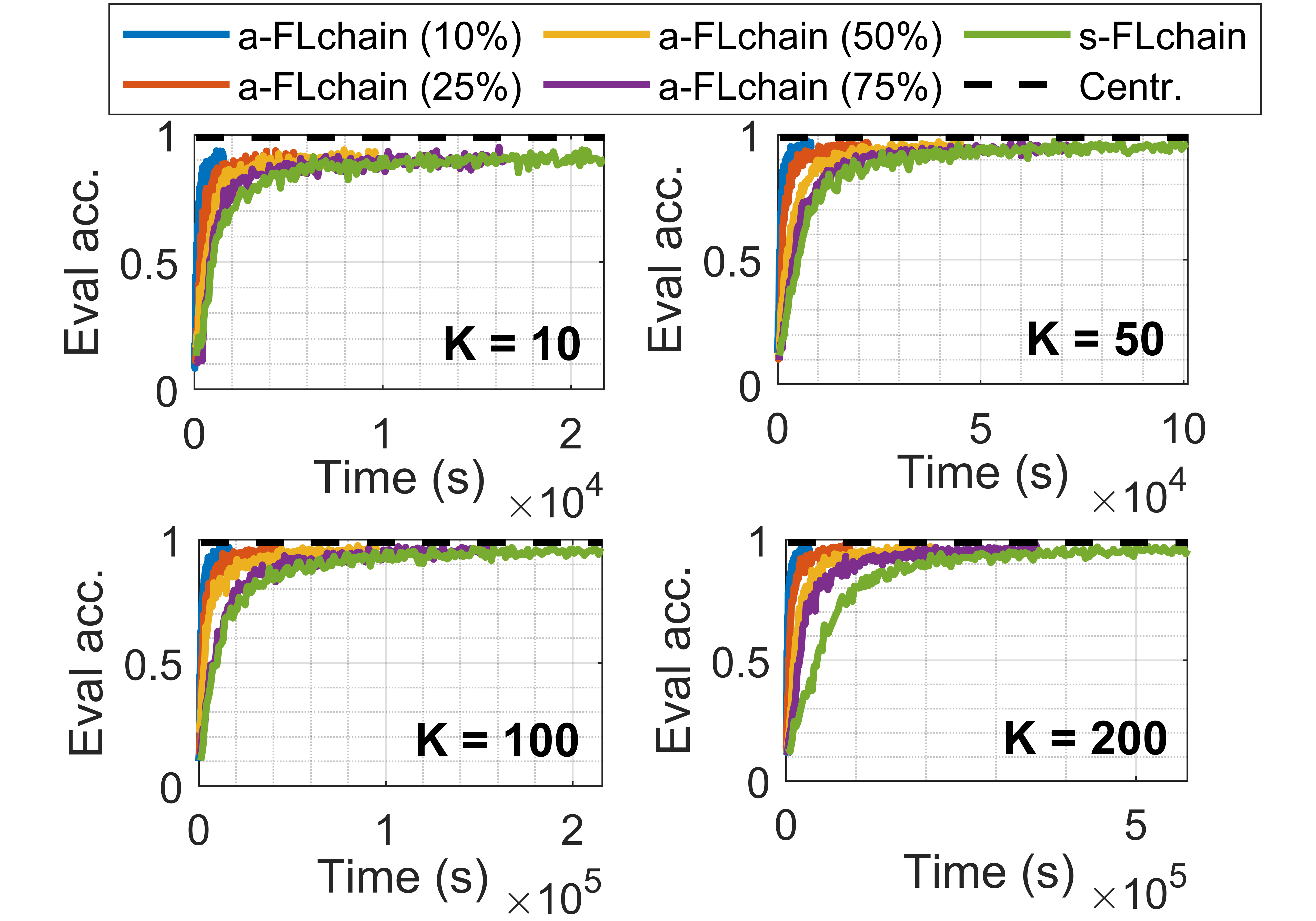}\label{fig:temporal_accuracy_emnist_cnn}}
    \subfigure[]{\includegraphics[width=.45\textwidth]{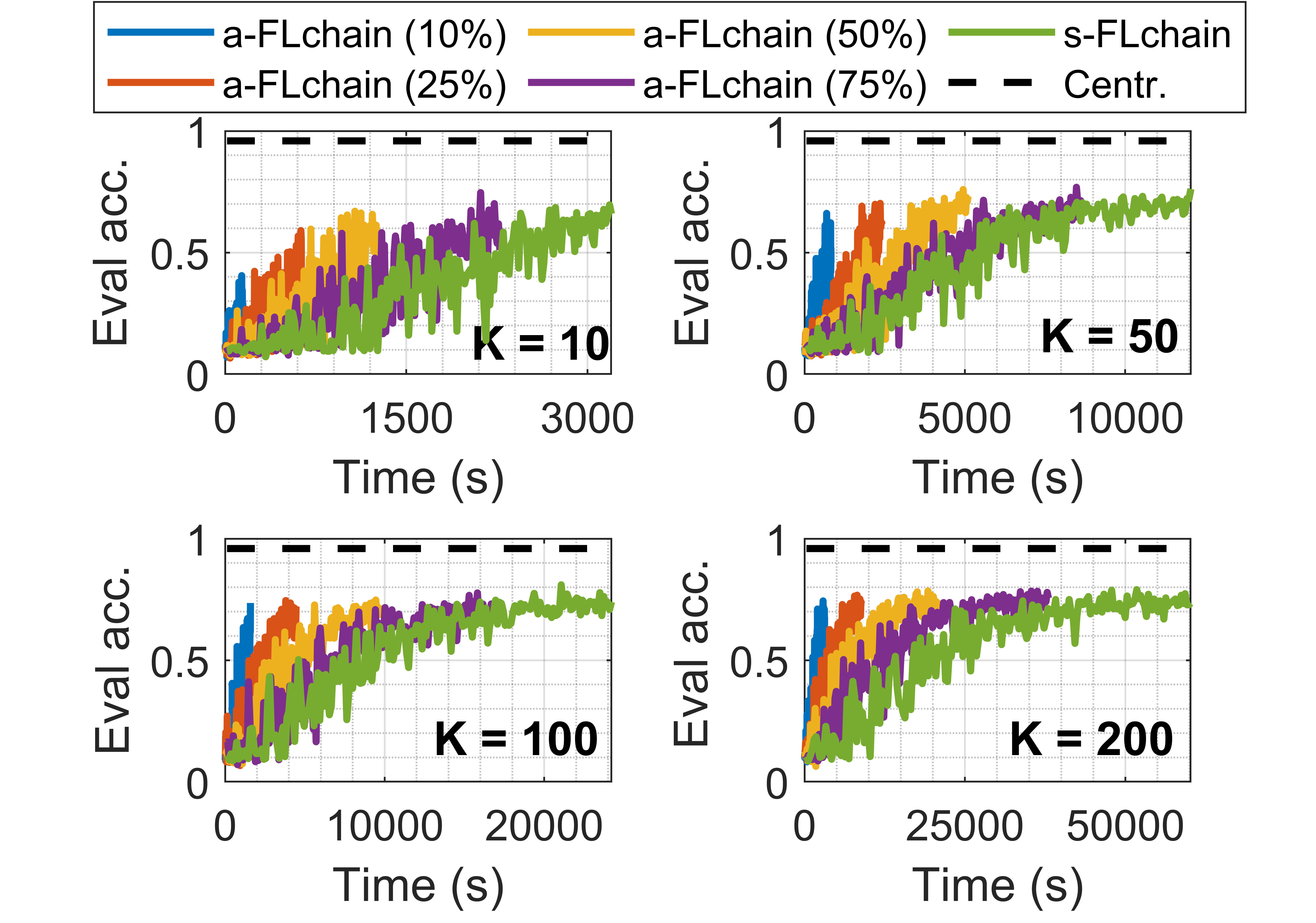}\label{fig:temporal_accuracy_emnist_fnn_noniid}}
	\subfigure[]{\includegraphics[width=.45\textwidth]{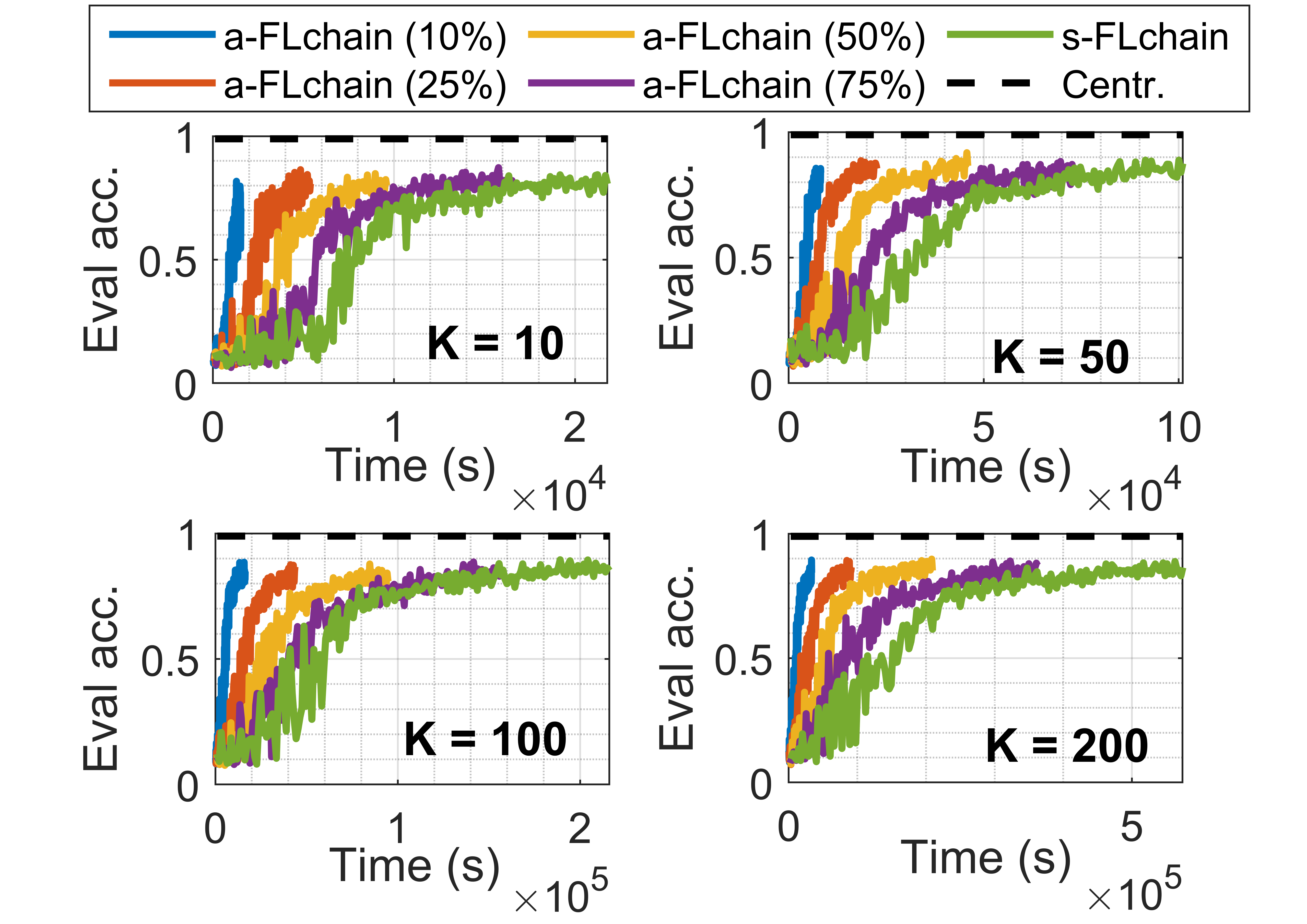}\label{fig:temporal_accuracy_emnist_cnn_noniid}}
	\caption{Evaluation accuracy obtained by s-FLchain and a-FLchain on the EMNIST dataset after $T=200$ rounds, for different number of clients $K$ and client sub-sampling percentages~$\Upsilon$: (a) FNN (IID), (b) CNN (IID), (c) FNN (non-IID), (d) CNN (non-IID). The centralized counterpart of each method is shown by a black dashed line.}
	\label{fig:temporal_evolution}
\end{figure*}
The analyzed models have been shown to achieve a higher accuracy as $\Upsilon$ increases, being s-FLchain the best-performing method. Let us now focus on the blockchain latency and the temporal performance obtained by each model throughout the considered $R=200$ FL rounds. For that, we use the model provided in Section~\ref{section:delay_bc} to compute the iteration time in each of the applied mechanisms for each block size. Taking this into account, Fig.~\ref{fig:temporal_evolution} illustrates the evaluation accuracy and loss with respect to the time each method takes to complete the 200 rounds, for both IID and non-IID settings.

As shown in Fig.~\ref{fig:temporal_evolution}, the longer the block size (determined by the total number of users, $K$, and the percentage of users participating in each round, $\Upsilon$), the higher the training time for completing the considered 200 FL rounds. This makes a-FLchain more efficient than s-FLchain in terms of iteration time (especially for low $\Upsilon$ values), provided that the synchronized method requires a lot of time to include all the local updates to a single block. However, there is a trade-off between accuracy and training time. Regarding the models used, we observe that the CNN achieves higher accuracy in all the cases than FNN (see Fig.~\ref{fig:temporal_accuracy_emnist_fnn} and Fig.~\ref{fig:temporal_accuracy_emnist_cnn}), but at the cost of requiring much more time for exchanging the models through the blockchain. Interestingly, the CNN is able to provide acceptable accuracy in the non-IID setting (see Fig.~\ref{fig:temporal_accuracy_emnist_cnn_noniid}), even for small block sizes (e.g., $K=10$ and $\Upsilon=25\%$). This is not the case for the FNN model (see Fig.~\ref{fig:temporal_accuracy_emnist_fnn_noniid}), which performs poorly in the non-IID setting when the block size is small.

Next, to further analyze the accuracy versus training time trade-off, Table~\ref{tab:acuracy_per_second} provides the results on the training efficiency achieved by each method in every scenario. The training efficiency is computed as the average accuracy achieved by each method divided by the iteration time.

\begin{table*}[ht!]
\centering
\caption{Average accuracy per second achieved by the FNN and CNN models in EMNIST IID (EMNIST non-IID).}
\label{tab:acuracy_per_second}
\resizebox{\textwidth}{!}{\begin{tabular}{|cc|ccccc|}
\hline
\multicolumn{2}{|c|}{\multirow{2}{*}{}} & \multicolumn{5}{c|}{\textbf{Block size}} \\ \cline{3-7} 
\multicolumn{2}{|c|}{} & \multicolumn{1}{c|}{\textbf{$\Upsilon=10\%$}} & \multicolumn{1}{c|}{\textbf{$\Upsilon=25\%$}} & \multicolumn{1}{c|}{\textbf{$\Upsilon=50\%$}} & \multicolumn{1}{c|}{\textbf{$\Upsilon=75\%$}} & \textbf{$\Upsilon=100\%$} \\ \hline
\multicolumn{1}{|c|}{\multirow{2}{*}{\textbf{K = 10}}} & \textit{FNN} & \multicolumn{1}{c|}{$86\cdot 10^{-3}$ ($26.5\cdot 10^{-2}$)} & \multicolumn{1}{c|}{$21.3\cdot 10^{-2}$ ($7.1\cdot 10^{-2}$)} & \multicolumn{1}{c|}{$11.2\cdot 10^{-2}$ ($4.9\cdot 10^{-2}$)} & \multicolumn{1}{c|}{$6.2\cdot 10^{-2}$ ($2.9\cdot 10^{-2}$)} & $4.4\cdot 10^{-2}$ ($2.2\cdot 10^{-2}$) \\ \cline{2-7} 
\multicolumn{1}{|c|}{} & \textit{CNN} & \multicolumn{1}{c|}{$10.89\cdot 10^{-2}$ ($48.9\cdot 10^{-2}$)} & \multicolumn{1}{c|}{$26\cdot 10^{-2}$ ($15.5\cdot 10^{-2}$)} & \multicolumn{1}{c|}{$13.2\cdot 10^{-2}$ ($8.1\cdot 10^{-2}$)} & \multicolumn{1}{c|}{$7.4\cdot 10^{-2}$ ($4.9\cdot 10^{-2}$)} & $5.3\cdot 10^{-2}$ ($3.4\cdot 10^{-2}$) \\ \hline
\multicolumn{1}{|c|}{\multirow{2}{*}{\textbf{K = 50}}} & \textit{FNN} & \multicolumn{1}{c|}{$15.3\cdot 10^{-2}$ ($4.7\cdot 10^{-2}$)} & \multicolumn{1}{c|}{$5.6\cdot 10^{-2}$ ($1.9\cdot 10^{-2}$)} & \multicolumn{1}{c|}{$2.8\cdot 10^{-2}$ ($1.2\cdot 10^{-2}$)} & \multicolumn{1}{c|}{$1.6\cdot 10^{-2}$ ($8\cdot 10^{-3}$)} & $1.2\cdot 10^{-2}$ ($6\cdot 10^{-3}$) \\ \cline{2-7} 
\multicolumn{1}{|c|}{} & \textit{CNN} & \multicolumn{1}{c|}{$19\cdot 10^{-2}$ ($8.7\cdot 10^{-2}$)} & \multicolumn{1}{c|}{$6.8\cdot 10^{-2}$ ($4\cdot 10^{-2}$)} & \multicolumn{1}{c|}{$3.3\cdot 10^{-2}$ ($2\cdot 10^{-2}$)} & \multicolumn{1}{c|}{$1.9\cdot 10^{-2}$ ($1.3\cdot 10^{-2}$)} & $1.4\cdot 10^{-2}$ ($9\cdot 10^{-3}$) \\ \hline
\multicolumn{1}{|c|}{\multirow{2}{*}{\textbf{K = 100}}} & \textit{FNN} & \multicolumn{1}{c|}{$8\cdot 10^{-2}$ ($2.5\cdot 10^{-2}$)} & \multicolumn{1}{c|}{$3\cdot 10^{-2}$ ($1\cdot 10^{-2}$)} & \multicolumn{1}{c|}{$1.4\cdot 10^{-2}$ ($6\cdot 10^{-3}$)} & \multicolumn{1}{c|}{$8\cdot 10^{-3}$ ($4\cdot 10^{-3}$)} & $6\cdot 10^{-3}$ ($3\cdot 10^{-3}$) \\ \cline{2-7} 
\multicolumn{1}{|c|}{} & \textit{CNN} & \multicolumn{1}{c|}{$10.1\cdot 10^{-2}$ ($4.5\cdot 10^{-2}$)} & \multicolumn{1}{c|}{$3.7\cdot 10^{-2}$ ($2.2\cdot 10^{-2}$)} & \multicolumn{1}{c|}{$1.7\cdot 10^{-2}$ ($1\cdot 10^{-2}$)} & \multicolumn{1}{c|}{$1\cdot 10^{-2}$ ($6\cdot 10^{-3}$)} & $7\cdot 10^{-3}$ ($5\cdot 10^{-3}$) \\ \hline
\multicolumn{1}{|c|}{\multirow{2}{*}{\textbf{K = 200}}} & \textit{FNN} & \multicolumn{1}{c|}{$4\cdot 10^{-2}$ ($1.2\cdot 10^{-2}$)} & \multicolumn{1}{c|}{$1.5\cdot 10^{-2}$ ($5\cdot 10^{-3}$)} & \multicolumn{1}{c|}{$7\cdot 10^{-3}$ ($3\cdot 10^{-3}$)} & \multicolumn{1}{c|}{$4\cdot 10^{-3}$ ($1\cdot 10^{-3}$)} & $2\cdot 10^{-3}$ ($1\cdot 10^{-3}$) \\ \cline{2-7} 
\multicolumn{1}{|c|}{} & \textit{CNN} & \multicolumn{1}{c|}{$5.1\cdot 10^{-2}$ ($2.3\cdot 10^{-2}$)} & \multicolumn{1}{c|}{$1.9\cdot 10^{-2}$ ($1.1\cdot 10^{-2}$)} & \multicolumn{1}{c|}{$8\cdot 10^{-3}$ ($5\cdot 10^{-3}$)} & \multicolumn{1}{c|}{$4\cdot 10^{-3}$ ($3\cdot 10^{-3}$)} & $3\cdot 10^{-3}$ ($2\cdot 10^{-3}$) \\ \hline
\end{tabular}}
\end{table*}

Table~\ref{tab:acuracy_per_second} highlights the efficiency of the asynchronous mechanism, provided that efficiency decreases as $K$ and $\Upsilon$ increase. For instance, the highest achieved efficiency is obtained by the FNN in the scenario where a-FLchain with $\Upsilon=10\%$ is applied for $K=10$ users. This is an important conclusion that suggests that a-FLchain is an appealing solution when dealing with large distributed datasets in FL. In those cases, maintaining a strict synchronization among FL clients is counterproductive in terms of efficiency. However, the accuracy-time trade-off must be carefully considered when selecting a model and the type of blockchain to be used. Regarding the application of more complex models in FLchain, the CNN is shown to be more efficient than the FNN in a more challenging situation like the non-IID setting, where the simplicity of the FNN fails at capturing insightful patterns from data.

\begin{figure}[ht!]
	\centering
	\includegraphics[width=0.8\linewidth]{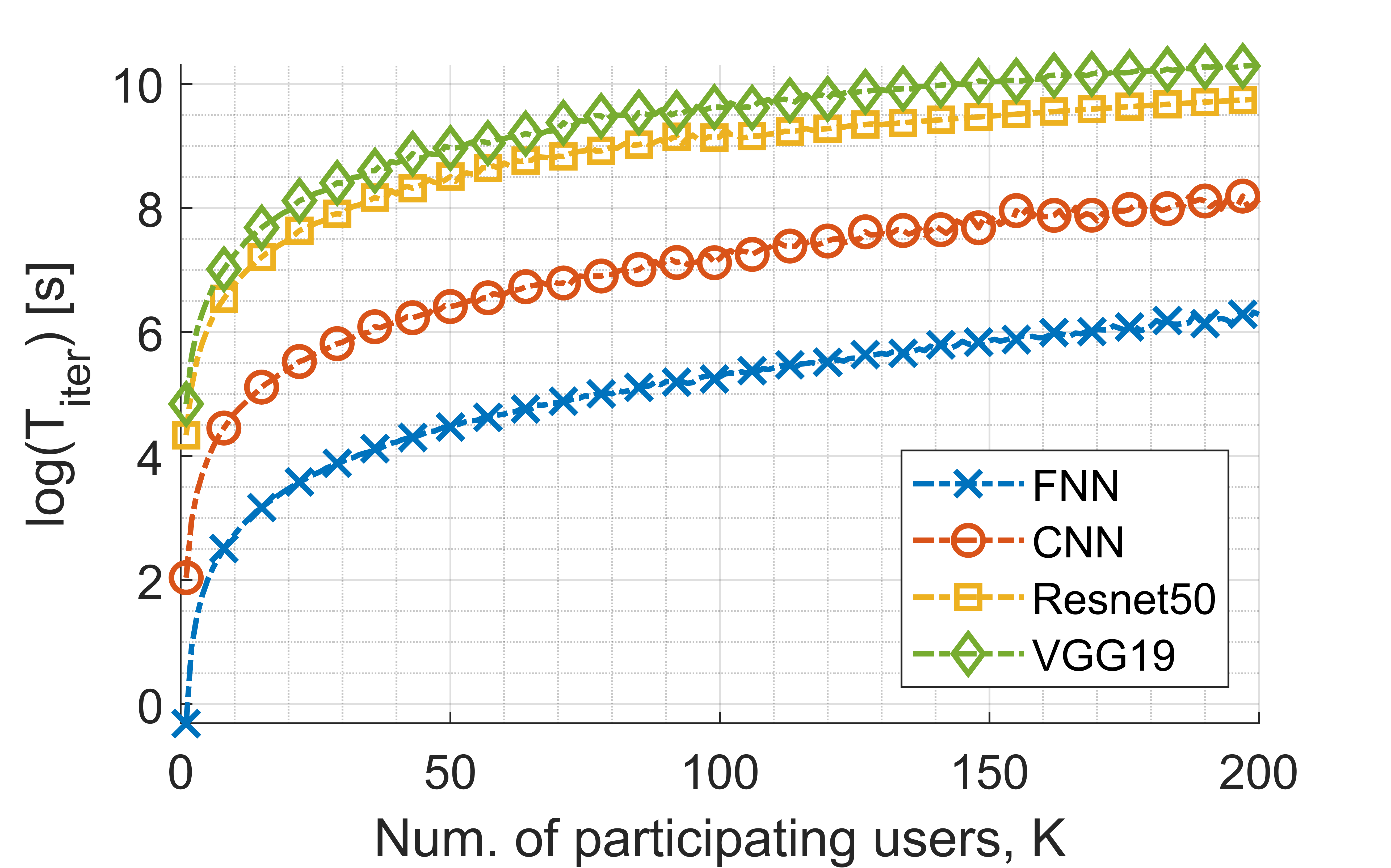}
	\caption{Comparison of the iteration delays achieved by different ML models in the blockchained FL setting.}	\label{fig:plot_delay_per_iteration_models}
\end{figure}

Despite more complex models can provide higher accuracy than simpler ones, as shown with the examples of FNN and CNN models, their size may inflict very high delays to the training procedure. In our evaluation, the most complex considered model was the CNN, which has a size of 4.749~MB. Nevertheless, much more complex deep learning models are used nowadays to solve more challenging problems (e.g., CIFAR-100~\cite{krizhevsky2009learning}). Two prominent examples of very deep models are residual neural networks (ResNet) and visual geometry group (VGG)~\cite{simonyan2014very}. The Resnet50 and VGG19 implementations use 23,792,612 and 39,316,644 parameters, respectively, thus requiring 47.58 MB and 78.63~MB of storage. These values throw into question the feasible implementation of very complex models in federated settings, as they entail a huge communication overhead. Figure~\ref{fig:plot_delay_per_iteration_models} shows the FL iteration delay of each of the abovementioned ML models for different numbers of participating users in FL. For the sake of illustration, the iteration time is shown on a logarithmic scale.

As shown, deep learning models result in very high delays when trained through a blockchained FL solution, being VGG19 up to four levels of magnitude with respect to FNN. The high differences are due to the big overheads experienced in the distributed blockchain system, which include phenomena like forks, whereby the confirmation time increases exponentially with the size of the exchanged transactions.

%%%%%%%%%%%%%%%%%%%%%%%%%%%%
%% CONCLUSIONS
%%%%%%%%%%%%%%%%%%%%%%%%%%%%
\section{Conclusions}
\label{section:conclusions}
This paper has addressed an emerging topic regarding distributed federated optimization and has proposed the usage of blockchain technology to realize it in a secure, transparent, and reliable manner. To assess the feasibility of this blockchain-enabled solution, a blockchain latency framework has been developed, including a novel queuing model suitable to blockchain technology. Based on this framework, synchronous and asynchronous methods (s-FLchain and a-FLchain) are evaluated in terms of accuracy and latency. While the synchronous version of models like FNN and CNN allow obtaining a high accuracy for both IID and non-IID datasets, their iteration time becomes very high as the number of participating users increases. This prevents the adoption of this kind of mechanism in massive deployments, thus lacking scalability. In contrast, the asynchronous FL optimization solution keeps communication overheads low, as it allows for reducing the number of local updates included in each iteration. However, the accuracy of asynchronous models is tied to the contributions made by individual FL devices, thus revealing the accuracy versus training time trade-off. In this regard, the development of high-performing lightweight ML models becomes of utmost importance to truly enable the distributed learning paradigm.

Future research directions include innovative mechanisms and architectural solutions for FLchain, able to address the challenges raised by either blockchain implementations (e.g., capacity, latency) or the device and data heterogeneity in asynchronous training. Another remarkable topic for future research is on the evaluation of the security properties of the FLchain solution. Blockchain security is a broad topic that involves many operations (e.g., cryptography of users' identities, decentralized consensus) and its application to FL can satisfy multiple purposes (e.g., validation of models, users' reputation, mitigating poisoning attacks). Furthermore, depending on the adopted blockchain solution, different security properties would be granted.

\ifCLASSOPTIONcompsoc
  \section*{Acknowledgments}
\else
  \section*{Acknowledgment}
\fi

This work was funded by the IN CERCA grant from the Secretaria d'Universitats i Recerca del departament d'Empresa i Coneixement de la Generalitat de Catalunya, and partially from the Spanish grant PID2020-113832RB-C22(ORIGIN)/MCIN/AEI/10.13039/5011000110, by the grant CHIST-ERA-20-SICT-004, by the Spanish grant PCI2021-122043-2A/AEI/10.13039/501100011033. 

\ifCLASSOPTIONcaptionsoff
  \newpage
\fi

\bibliographystyle{IEEEtran}
\bibliography{bib}

\begin{thebibliography}{10}
\providecommand{\url}[1]{#1}
\csname url@samestyle\endcsname
\providecommand{\newblock}{\relax}
\providecommand{\bibinfo}[2]{#2}
\providecommand{\BIBentrySTDinterwordspacing}{\spaceskip=0pt\relax}
\providecommand{\BIBentryALTinterwordstretchfactor}{4}
\providecommand{\BIBentryALTinterwordspacing}{\spaceskip=\fontdimen2\font plus
\BIBentryALTinterwordstretchfactor\fontdimen3\font minus
  \fontdimen4\font\relax}
\providecommand{\BIBforeignlanguage}[2]{{%
\expandafter\ifx\csname l@#1\endcsname\relax
\typeout{** WARNING: IEEEtran.bst: No hyphenation pattern has been}%
\typeout{** loaded for the language `#1'. Using the pattern for}%
\typeout{** the default language instead.}%
\else
\language=\csname l@#1\endcsname
\fi
#2}}
\providecommand{\BIBdecl}{\relax}
\BIBdecl

\bibitem{konevcny2016federated}
J.~Kone{\v{c}}n{\`y}, H.~B. McMahan, F.~X. Yu, P.~Richt{\'a}rik, A.~T. Suresh,
  and D.~Bacon, ``Federated learning: Strategies for improving communication
  efficiency,'' \emph{arXiv preprint arXiv:1610.05492}, 2016.

\bibitem{chen2020fedhealth}
Y.~Chen, X.~Qin, J.~Wang, C.~Yu, and W.~Gao, ``Fedhealth: A federated transfer
  learning framework for wearable healthcare,'' \emph{IEEE Intelligent
  Systems}, vol.~35, no.~4, pp. 83--93, 2020.

\bibitem{nguyen2021federated2}
D.~C. Nguyen, M.~Ding, P.~N. Pathirana, A.~Seneviratne, and A.~Y. Zomaya,
  ``Federated learning for covid-19 detection with generative adversarial
  networks in edge cloud computing,'' \emph{IEEE Internet of Things Journal},
  2021.

\bibitem{lim2020federated}
W.~Y.~B. Lim, N.~C. Luong, D.~T. Hoang, Y.~Jiao, Y.-C. Liang, Q.~Yang,
  D.~Niyato, and C.~Miao, ``Federated learning in mobile edge networks: A
  comprehensive survey,'' \emph{IEEE Comm. Surveys \& Tutorials}, vol.~22,
  no.~3, pp. 2031--2063, 2020.

\bibitem{wilhelmi2020flexible}
F.~Wilhelmi, S.~Barrachina-Munoz, B.~Bellalta, C.~Cano, A.~Jonsson, and V.~Ram,
  ``A flexible machine-learning-aware architecture for future wlans,''
  \emph{IEEE Comm. Mag.}, vol.~58, no.~3, pp. 25--31, 2020.

\bibitem{niknam2020federated}
S.~Niknam, H.~S. Dhillon, and J.~H. Reed, ``Federated learning for wireless
  communications: Motivation, opportunities, and challenges,'' \emph{IEEE Comm.
  Mag.}, vol.~58, no.~6, pp. 46--51, 2020.

\bibitem{du2020federated}
Z.~Du, C.~Wu, T.~Yoshinaga, K.-L.~A. Yau, Y.~Ji, and J.~Li, ``Federated
  learning for vehicular internet of things: Recent advances and open issues,''
  \emph{IEEE Open Journal of the Computer Society}, vol.~1, pp. 45--61, 2020.

\bibitem{brik2020federated}
B.~Brik, A.~Ksentini, and M.~Bouaziz, ``Federated learning for uavs-enabled
  wireless networks: Use cases, challenges, and open problems,'' \emph{IEEE
  Access}, vol.~8, pp. 53\,841--53\,849, 2020.

\bibitem{wang2019adaptive}
S.~Wang, T.~Tuor, T.~Salonidis, K.~K. Leung, C.~Makaya, T.~He, and K.~Chan,
  ``Adaptive federated learning in resource constrained edge computing
  systems,'' \emph{IEEE JSAC}, vol.~37, no.~6, pp. 1205--1221, 2019.

\bibitem{mills2019communication}
J.~Mills, J.~Hu, and G.~Min, ``Communication-efficient federated learning for
  wireless edge intelligence in iot,'' \emph{IEEE Internet Things J.}, vol.~7,
  no.~7, pp. 5986--5994, 2019.

\bibitem{nakamoto2008bitcoin}
S.~Nakamoto, ``{Bitcoin: A Peer-to-Peer electronic cash system},'' Tech. Rep.,
  2008.

\bibitem{li2020federated2}
T.~Li, A.~K. Sahu, A.~Talwalkar, and V.~Smith, ``Federated learning:
  Challenges, methods, and future directions,'' \emph{IEEE Signal Processing
  Magazine}, vol.~37, no.~3, pp. 50--60, 2020.

\bibitem{sprague2018asynchronous}
M.~R. Sprague, A.~Jalalirad, M.~Scavuzzo, C.~Capota, M.~Neun, L.~Do, and
  M.~Kopp, ``Asynchronous federated learning for geospatial applications,'' in
  \emph{Joint European Conference on Machine Learning and Knowledge Discovery
  in Databases}.\hskip 1em plus 0.5em minus 0.4em\relax Springer, 2018, pp.
  21--28.

\bibitem{feyzmahdavian2016asynchronous}
H.~R. Feyzmahdavian, A.~Aytekin, and M.~Johansson, ``An asynchronous mini-batch
  algorithm for regularized stochastic optimization,'' \emph{IEEE Trans. Autom.
  Control}, vol.~61, no.~12, pp. 3740--3754, 2016.

\bibitem{xu2021bafl}
C.~Xu, Y.~Qu, P.~W. Eklund, Y.~Xiang, and L.~Gao, ``Bafl: an efficient
  blockchain-based asynchronous federated learning framework,'' in \emph{2021
  IEEE Symposium on Computers and Communications (ISCC)}.\hskip 1em plus 0.5em
  minus 0.4em\relax IEEE, 2021, pp. 1--6.

\bibitem{nguyen2021bedgehealth}
D.~C. Nguyen, P.~N. Pathirana, M.~Ding, and A.~Seneviratne, ``Bedgehealth: A
  decentralized architecture for edge-based iomt networks using blockchain,''
  \emph{IEEE Internet of Things Journal}, vol.~8, no.~14, pp. 11\,743--11\,757,
  2021.

\bibitem{fang2020local}
M.~Fang, X.~Cao, J.~Jia, and N.~Gong, ``Local model poisoning attacks to
  $\{$Byzantine-Robust$\}$ federated learning,'' in \emph{29th USENIX Security
  Symposium (USENIX Security 20)}, 2020, pp. 1605--1622.

\bibitem{mothukuri2021survey}
V.~Mothukuri, R.~M. Parizi, S.~Pouriyeh, Y.~Huang, A.~Dehghantanha, and
  G.~Srivastava, ``A survey on security and privacy of federated learning,''
  \emph{Future Generation Computer Systems}, vol. 115, pp. 619--640, 2021.

\bibitem{nguyen2019blockchain}
D.~C. Nguyen, P.~N. Pathirana, M.~Ding, and A.~Seneviratne, ``Blockchain for
  secure ehrs sharing of mobile cloud based e-health systems,'' \emph{IEEE
  access}, vol.~7, pp. 66\,792--66\,806, 2019.

\bibitem{zhan2020learning}
Y.~Zhan, P.~Li, Z.~Qu, D.~Zeng, and S.~Guo, ``A learning-based incentive
  mechanism for federated learning,'' \emph{IEEE Internet of Things Journal},
  vol.~7, no.~7, pp. 6360--6368, 2020.

\bibitem{somy2019ownership}
N.~B. Somy, K.~Kannan, V.~Arya, S.~Hans, A.~Singh, P.~Lohia, and S.~Mehta,
  ``Ownership preserving ai market places using blockchain,'' in \emph{2019
  IEEE International Conference on Blockchain (Blockchain)}.\hskip 1em plus
  0.5em minus 0.4em\relax IEEE, 2019, pp. 156--165.

\bibitem{tu2022incentive}
X.~Tu, K.~Zhu, N.~C. Luong, D.~Niyato, Y.~Zhang, and J.~Li, ``Incentive
  mechanisms for federated learning: From economic and game theoretic
  perspective,'' \emph{IEEE Transactions on Cognitive Communications and
  Networking}, 2022.

\bibitem{kairouz2021advances}
P.~Kairouz, H.~B. McMahan, B.~Avent, A.~Bellet, M.~Bennis, A.~N. Bhagoji,
  K.~Bonawitz, Z.~Charles, G.~Cormode, R.~Cummings \emph{et~al.}, ``Advances
  and open problems in federated learning,'' \emph{Foundations and
  Trends{\textregistered} in Machine Learning}, vol.~14, no. 1--2, pp. 1--210,
  2021.

\bibitem{nguyen2021federated}
D.~C. Nguyen, M.~Ding, Q.-V. Pham, P.~N. Pathirana, L.~B. Le, A.~Seneviratne,
  J.~Li, D.~Niyato, and H.~V. Poor, ``Federated learning meets blockchain in
  edge computing: Opportunities and challenges,'' \emph{IEEE Internet Things
  J.}, 2021.

\bibitem{li2020survey}
X.~Li, P.~Jiang, T.~Chen, X.~Luo, and Q.~Wen, ``A survey on the security of
  blockchain systems,'' \emph{Future Generation Computer Systems}, vol. 107,
  pp. 841--853, 2020.

\bibitem{zhao2020privacy}
Y.~Zhao, J.~Zhao, L.~Jiang, R.~Tan, D.~Niyato, Z.~Li, L.~Lyu, and Y.~Liu,
  ``Privacy-preserving blockchain-based federated learning for iot devices,''
  \emph{IEEE Internet of Things Journal}, vol.~8, no.~3, pp. 1817--1829, 2020.

\bibitem{majeed2019flchain}
U.~Majeed and C.~S. Hong, ``Flchain: Federated learning via mec-enabled
  blockchain network,'' in \emph{2019 APNOMS}.\hskip 1em plus 0.5em minus
  0.4em\relax IEEE, 2019, pp. 1--4.

\bibitem{bao2019flchain}
X.~Bao, C.~Su, Y.~Xiong, W.~Huang, and Y.~Hu, ``Flchain: A blockchain for
  auditable federated learning with trust and incentive,'' in \emph{2019
  BIGCOM}.\hskip 1em plus 0.5em minus 0.4em\relax IEEE, 2019, pp. 151--159.

\bibitem{zheng2018blockchain}
Z.~Zheng, S.~Xie, H.-N. Dai, X.~Chen, and H.~Wang, ``Blockchain challenges and
  opportunities: A survey,'' \emph{Int. Journal of Web and Grid Services},
  vol.~14, no.~4, pp. 352--375, 2018.

\bibitem{wilhelmi2021discrete}
F.~Wilhelmi and L.~Giupponi, ``Discrete-time analysis of wireless blockchain
  networks,'' \emph{arXiv preprint arXiv:2104.05586}, 2021.

\bibitem{cohen_afshar_tapson_schaik_2017}
G.~Cohen, S.~Afshar, J.~Tapson, and A.~V. Schaik, ``Emnist: Extending mnist to
  handwritten letters,'' \emph{2017 International Joint Conference on Neural
  Networks (IJCNN)}, 2017.

\bibitem{lalitha2018fully}
A.~Lalitha, S.~Shekhar, T.~Javidi, and F.~Koushanfar, ``Fully decentralized
  federated learning,'' in \emph{NeurIPS}, 2018.

\bibitem{hegedHus2021decentralized}
I.~Heged{\H{u}}s, G.~Danner, and M.~Jelasity, ``Decentralized learning works:
  An empirical comparison of gossip learning and federated learning,''
  \emph{Journal of Parallel and Distributed Computing}, vol. 148, pp. 109--124,
  2021.

\bibitem{liu2020secure}
Y.~Liu, J.~Peng, J.~Kang, A.~M. Iliyasu, D.~Niyato, and A.~A. Abd El-Latif, ``A
  secure federated learning framework for 5g networks,'' \emph{IEEE Wireless
  Commun.}, vol.~27, no.~4, pp. 24--31, 2020.

\bibitem{hou2021systematic}
D.~Hou, J.~Zhang, K.~L. Man, J.~Ma, and Z.~Peng, ``A systematic literature
  review of blockchain-based federated learning: Architectures, applications
  and issues,'' in \emph{2021 ICTC}.\hskip 1em plus 0.5em minus 0.4em\relax
  IEEE, 2021, pp. 302--307.

\bibitem{lu2020blockchain2}
Y.~Lu, X.~Huang, K.~Zhang, S.~Maharjan, and Y.~Zhang, ``Blockchain and
  federated learning for 5g beyond,'' \emph{IEEE Network}, vol.~35, no.~1, pp.
  219--225, 2020.

\bibitem{lu2019blockchain}
Y.~Lu, X.~Huang, Y.~Dai, S.~Maharjan, and Y.~Zhang, ``Blockchain and federated
  learning for privacy-preserved data sharing in industrial iot,'' \emph{IEEE
  Trans. Ind. Informat.}, vol.~16, no.~6, pp. 4177--4186, 2019.

\bibitem{qu2020blockchained}
Y.~Qu, S.~R. Pokhrel, S.~Garg, L.~Gao, and Y.~Xiang, ``A blockchained federated
  learning framework for cognitive computing in industry 4.0 networks,''
  \emph{IEEE Trans. Ind. Informat.}, vol.~17, no.~4, pp. 2964--2973, 2020.

\bibitem{pokhrel2020federated}
S.~R. Pokhrel and J.~Choi, ``Federated learning with blockchain for autonomous
  vehicles: Analysis and design challenges,'' \emph{IEEE Trans. Commun.},
  vol.~68, no.~8, pp. 4734--4746, 2020.

\bibitem{qi2021privacy}
Y.~Qi, M.~S. Hossain, J.~Nie, and X.~Li, ``Privacy-preserving blockchain-based
  federated learning for traffic flow prediction,'' \emph{Future Generation
  Computer Systems}, vol. 117, pp. 328--337, 2021.

\bibitem{li2020federated}
T.~Li, A.~K. Sahu, M.~Zaheer, M.~Sanjabi, A.~Talwalkar, and V.~Smith,
  ``Federated optimization in heterogeneous networks,'' \emph{Proceedings of
  Machine Learning and Systems}, vol.~2, pp. 429--450, 2020.

\bibitem{wilhelmi2021performance}
F.~Wilhelmi and L.~Giupponi, ``On the performance of blockchain-enabled
  ran-as-a-service in beyond 5g networks,'' \emph{arXiv preprint
  arXiv:2105.14221}, 2021.

\bibitem{kim2019blockchained}
H.~Kim, J.~Park, M.~Bennis, and S.-L. Kim, ``Blockchained on-device federated
  learning,'' \emph{IEEE Comm. Lett.}, vol.~24, no.~6, pp. 1279--1283, 2019.

\bibitem{liu2021blockchain}
Y.~Liu, Y.~Qu, C.~Xu, Z.~Hao, and B.~Gu, ``Blockchain-enabled asynchronous
  federated learning in edge computing,'' \emph{Sensors}, vol.~21, no.~10, p.
  3335, 2021.

\bibitem{xie2019asynchronous}
C.~Xie, S.~Koyejo, and I.~Gupta, ``Asynchronous federated optimization,''
  \emph{arXiv preprint arXiv:1903.03934}, 2019.

\bibitem{xu2021asynchronous}
C.~Xu, Y.~Qu, Y.~Xiang, and L.~Gao, ``Asynchronous federated learning on
  heterogeneous devices: A survey,'' \emph{arXiv preprint arXiv:2109.04269},
  2021.

\bibitem{lian2018asynchronous}
X.~Lian, W.~Zhang, C.~Zhang, and J.~Liu, ``Asynchronous decentralized parallel
  stochastic gradient descent,'' in \emph{International Conference on Machine
  Learning}.\hskip 1em plus 0.5em minus 0.4em\relax PMLR, 2018, pp. 3043--3052.

\bibitem{lu2020blockchain}
Y.~Lu, X.~Huang, K.~Zhang, S.~Maharjan, and Y.~Zhang, ``Blockchain empowered
  asynchronous federated learning for secure data sharing in internet of
  vehicles,'' \emph{IEEE Trans. Veh. Technol.}, vol.~69, no.~4, pp. 4298--4311,
  2020.

\bibitem{feng2021blockchain}
L.~Feng, Y.~Zhao, S.~Guo, X.~Qiu, W.~Li, and P.~Yu, ``Blockchain-based
  asynchronous federated learning for internet of things,'' \emph{IEEE Trans.
  Comput.}, 2021.

\bibitem{kawase2017transaction}
Y.~Kawase and S.~Kasahara, ``Transaction-confirmation time for bitcoin: A
  queueing analytical approach to blockchain mechanism,'' in \emph{Int. Conf.
  on Queueing Theory and Network Applications}.\hskip 1em plus 0.5em minus
  0.4em\relax Springer, 2017, pp. 75--88.

\bibitem{kawase2018batch}
{Y. Kawase and S. Kasahara}, ``A batch-service queueing system with general
  input and its application to analysis of mining process for bitcoin
  blockchain,'' in \emph{2018 IEEE iThings and IEEE GreenCom and IEEE CPSCom
  and IEEE SmartData}.\hskip 1em plus 0.5em minus 0.4em\relax IEEE, 2018, pp.
  1440--1447.

\bibitem{li2018blockchain}
Q.-L. Li, J.-Y. Ma, and Y.-X. Chang, ``Blockchain queue theory,'' in \emph{Int.
  Conf. on Comput. Social Networks}.\hskip 1em plus 0.5em minus 0.4em\relax
  Springer, 2018, pp. 25--40.

\bibitem{geissler2019discrete}
S.~Geissler, T.~Prantl, S.~Lange, F.~Wamser, and T.~Hossfeld, ``Discrete-time
  analysis of the blockchain distributed ledger technology,'' in \emph{ITC
  31}.\hskip 1em plus 0.5em minus 0.4em\relax IEEE, 2019, pp. 130--137.

\bibitem{mcmahan2017communication}
B.~McMahan, E.~Moore, D.~Ramage, S.~Hampson, and B.~A. y~Arcas,
  ``Communication-efficient learning of deep networks from decentralized
  data,'' in \emph{Artificial intelligence and statistics}.\hskip 1em plus
  0.5em minus 0.4em\relax PMLR, 2017, pp. 1273--1282.

\bibitem{reddi2020adaptive}
S.~Reddi, Z.~Charles, M.~Zaheer, Z.~Garrett, K.~Rush, J.~Kone{\v{c}}n{\`y},
  S.~Kumar, and H.~B. McMahan, ``Adaptive federated optimization,'' \emph{arXiv
  preprint arXiv:2003.00295}, 2020.

\bibitem{chen2020convergence}
M.~Chen, H.~V. Poor, W.~Saad, and S.~Cui, ``Convergence time optimization for
  federated learning over wireless networks,'' \emph{IEEE Trans. Wireless
  Commun.}, vol.~20, no.~4, pp. 2457--2471, 2020.

\bibitem{li2018federated}
T.~Li, A.~K. Sahu, M.~Zaheer, M.~Sanjabi, A.~Talwalkar, and V.~Smith,
  ``Federated optimization in heterogeneous networks,'' \emph{arXiv preprint
  arXiv:1812.06127}, 2018.

\bibitem{chen2020asynchronous}
Y.~Chen, Y.~Ning, M.~Slawski, and H.~Rangwala, ``Asynchronous online federated
  learning for edge devices with non-iid data,'' in \emph{2020 IEEE Int. Conf.
  on Big Data}.\hskip 1em plus 0.5em minus 0.4em\relax IEEE, 2020, pp. 15--24.

\bibitem{saleh2021blockchain}
F.~Saleh, ``Blockchain without waste: Proof-of-stake,'' \emph{The Review of
  financial studies}, vol.~34, no.~3, pp. 1156--1190, 2021.

\bibitem{castro1999practical}
M.~Castro, B.~Liskov \emph{et~al.}, ``Practical byzantine fault tolerance,'' in
  \emph{OSDI}, vol.~99, no. 1999, 1999, pp. 173--186.

\bibitem{schwartz2014ripple}
D.~Schwartz, N.~Youngs, A.~Britto \emph{et~al.}, ``The ripple protocol
  consensus algorithm,'' \emph{Ripple Labs Inc White Paper}, vol.~5, no.~8, p.
  151, 2014.

\bibitem{ma2020federated}
C.~Ma, J.~Li, M.~Ding, L.~Shi, T.~Wang, Z.~Han, and H.~V. Poor, ``When
  federated learning meets blockchain: A new distributed learning paradigm,''
  \emph{arXiv preprint arXiv:2009.09338}, 2020.

\bibitem{nguyen2018survey}
G.-T. Nguyen and K.~Kim, ``A survey about consensus algorithms used in
  blockchain,'' \emph{Journal of Information processing systems}, vol.~14,
  no.~1, pp. 101--128, 2018.

\bibitem{decker2013information}
C.~Decker and R.~Wattenhofer, ``Information propagation in the bitcoin
  network,'' in \emph{IEEE P2P 2013 Proceedings}.\hskip 1em plus 0.5em minus
  0.4em\relax IEEE, 2013, pp. 1--10.

\bibitem{shortle2018fundamentals}
J.~F. Shortle, J.~M. Thompson, D.~Gross, and C.~M. Harris, \emph{Fundamentals
  of queueing theory}.\hskip 1em plus 0.5em minus 0.4em\relax John Wiley \&
  Sons, 2018, vol. 399.

\bibitem{he2016deep}
K.~He, X.~Zhang, S.~Ren, and J.~Sun, ``Deep residual learning for image
  recognition,'' in \emph{Proceedings of the IEEE conference on computer vision
  and pattern recognition}, 2016, pp. 770--778.

\bibitem{krizhevsky2009learning}
A.~Krizhevsky, G.~Hinton \emph{et~al.}, ``Learning multiple layers of features
  from tiny images,'' 2009.

\bibitem{simonyan2014very}
K.~Simonyan and A.~Zisserman, ``Very deep convolutional networks for
  large-scale image recognition,'' \emph{arXiv preprint arXiv:1409.1556}, 2014.

\end{thebibliography}

\vspace{-1cm}
\begin{IEEEbiography}[{\includegraphics[width=1in,height=1.25in,clip,keepaspectratio]{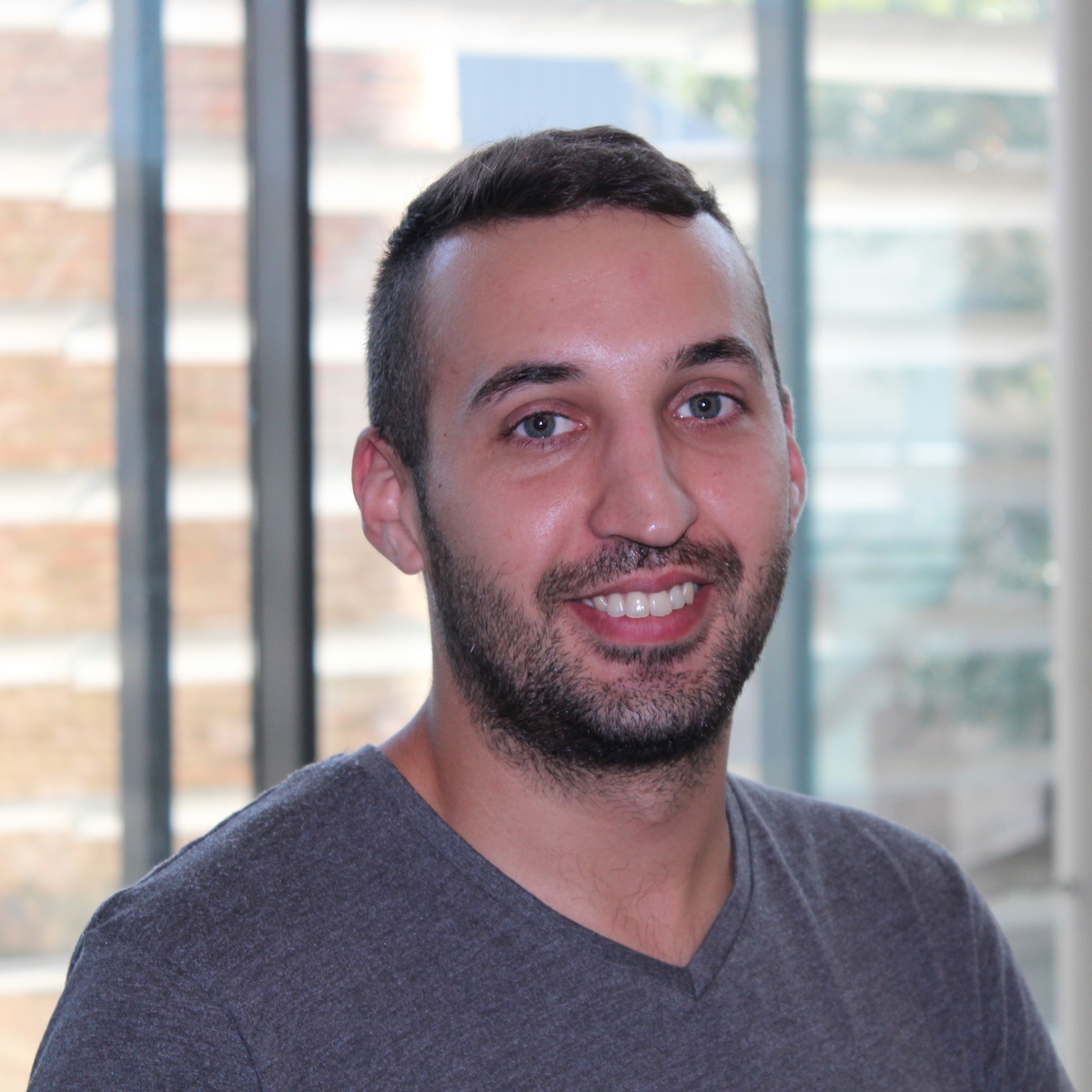}}]{Francesc Wilhelmi}
holds a Ph.D. in information and communication technologies (2020), from Universitat Pompeu Fabra (UPF). He is currently working as a postdoctoral researcher in the Mobile Networks department at Centre Tecnològic de Telecomunicacions de Catalunya (CTTC). 
\end{IEEEbiography}

\vspace{-1cm}
\begin{IEEEbiography}[{\includegraphics[width=1in,height=1.25in,clip,keepaspectratio]{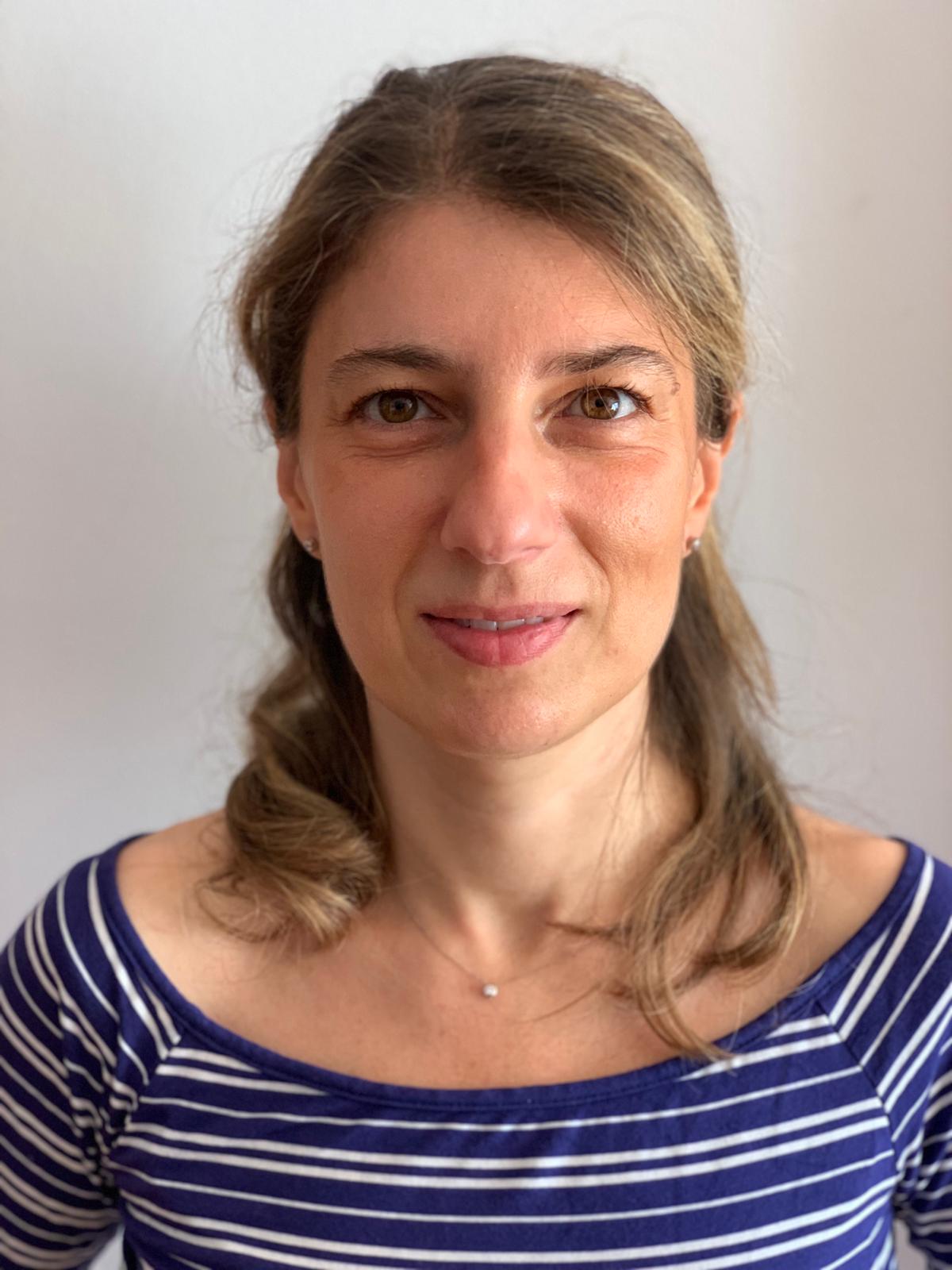}}]{Lorenza Giupponi} received her Ph.D. degree from UPC, Barcelona, Spain, in 2007. In 2003, she joined the Radio Communications Group, UPC, with a grant of the Spanish Ministry of  Education. From 2006 to 2007, she was an Assistant Professor with UPC. In September 2007, she joined CTTC, where she was a Research Director within the Mobile Networks Department. Between 2007 and 2011 she also was a member of the Executive Committee of CTTC, where she acted as the Director of Institutional Relations. She was a co-recipient of the IEEE CCNC 2010, the IEEE 3rd International Workshop on Indoor and Outdoor Femto Cells 2011, and the IEEE WCNC 2018 Best Paper Award. Between 2015 and 2011, she was a member of the Executive Committee of ns-3 Consortium in charge of coordinating 3GPP related developments. Since 2021 she is a Standardization Researcher in Ericsson.
\end{IEEEbiography}

\vspace{-1cm}
\begin{IEEEbiography}[{\includegraphics[width=1in,height=1.25in,clip,keepaspectratio]{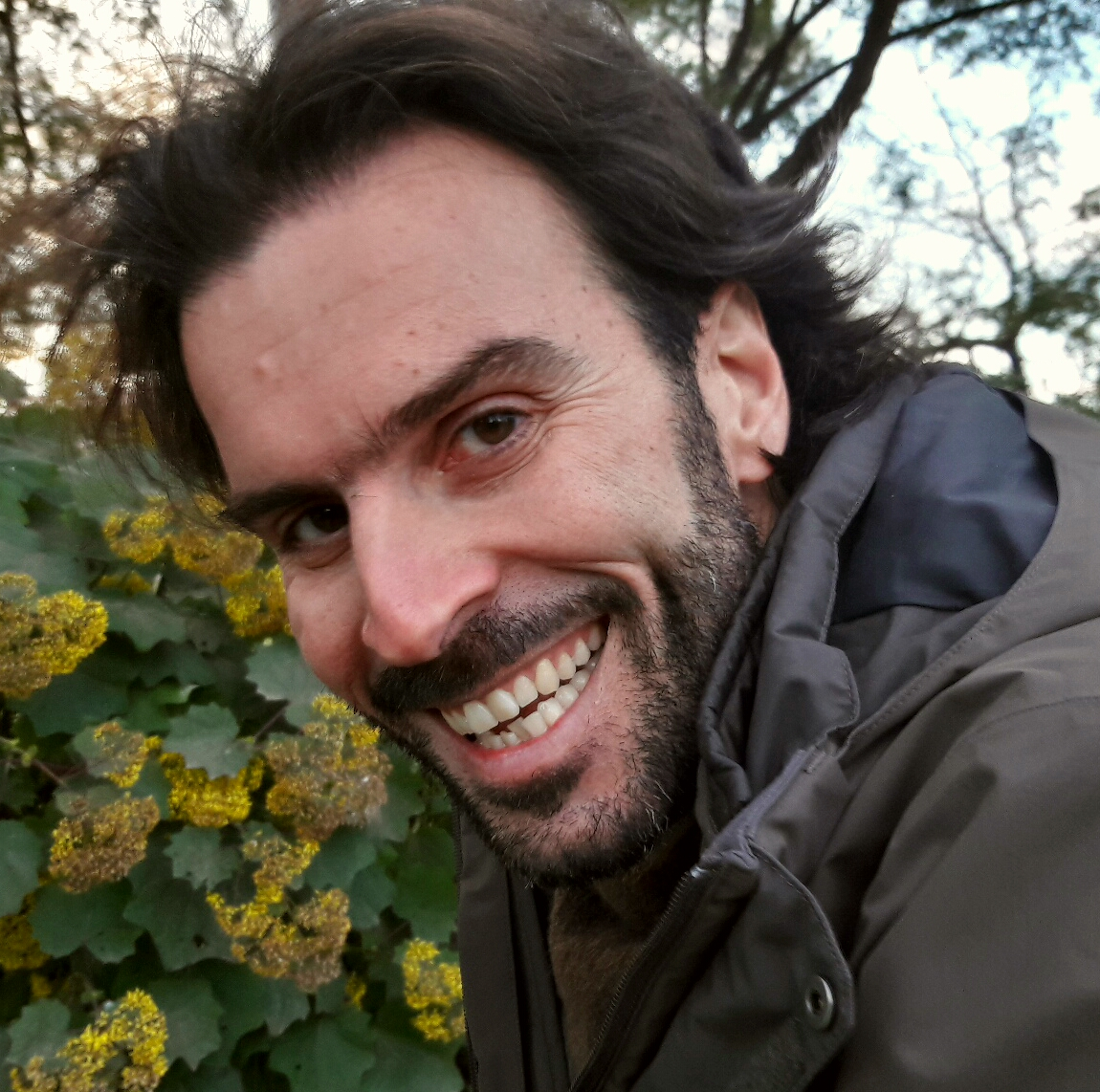}}]{Paolo Dini} received M.Sc. and Ph.D. from the Universit`a di Roma La Sapienza, in 2001 and 2005, respectively. He is currently a Senior Researcher with the Centre Tecnologic de Telecomunicacions de Catalunya (CTTC). His current research interests include sustainable networking and computing, distributed optimization and optimal control, machine learning, multi-agent systems and data analytics. His research activity is documented in almost 90 peer-reviewed scientific journals and international conference papers. He received two awards from the Cisco Silicon Valley Foundation for his research on heterogeneous mobile networks, in 2008 and 2011, respectively. He has been involved in more than 20 research and development projects. He is currently the Coordinator of CHIST-ERA SONATA project on sustainable computing and communication at the edge and the Scientific Coordinator of the EU H2020 MSCA Greenedge European Training Network on edge intelligence and sustainable computing. He serves as a TPC in many international conferences and workshops and as a reviewer for several scientific journals of the IEEE, Elsevier, ACM, Springer, Wiley.
\end{IEEEbiography}

\end{document}